\documentclass{article} 
\usepackage{iclr2016_conference}

\usepackage{times}
\usepackage{epsfig}
\usepackage{epstopdf}
\usepackage{ifpdf}
\usepackage{graphicx}
\usepackage{amsmath}
\usepackage{amssymb}
\usepackage{bbold}
\usepackage{array}
\usepackage{booktabs}
\usepackage{color}
\usepackage{subcaption}
\usepackage{tablefootnote}
\usepackage{hyperref}
\usepackage{url}
\usepackage{footnote}
\usepackage[hang]{footmisc}
\title{ParseNet: Looking Wider to See Better}

\author{Wei Liu\\
UNC Chapel Hill\\
\texttt{wliu@cs.unc.edu} \\
\And
Andrew Rabinovich\\
MagicLeap Inc.\\
\texttt{arabinovich@magicleap.com} \\
\And
Alexander C. Berg\\
UNC Chapel Hill\\
\texttt{aberg@cs.unc.edu}
}

\DeclareRobustCommand\onedot{\futurelet\@let@token\@onedot}
\def\@onedot{\ifx\@let@token.\else.\null\fi\xspace}

\def\etal{\emph{et al}\onedot}

\begin{document}

\maketitle

\begin{abstract}
We present a technique for adding global context to fully convolutional networks for semantic segmentation. The approach is simple, using the average feature for a layer to augment the features at each location. In addition, we study several idiosyncrasies of training, significantly increasing the performance of baseline networks (e.g. from FCN~\cite{long2014fully}). When we add our proposed global feature, and a technique for learning normalization parameters, accuracy increases consistently even over our improved versions of the baselines. Our proposed approach, ParseNet, achieves state-of-the-art performance on SiftFlow and PASCAL-Context with small additional computational cost over baselines, and near current state-of-the-art performance on PASCAL VOC 2012 semantic segmentation with a simple approach. Code is available at \url{https://github.com/weiliu89/caffe/tree/fcn} .
\end{abstract}

\section{Introduction}
Semantic segmentation, largely studied in the last 10 years, merges image segmentation with object recognition to produce per-pixel labeling of image content. The currently most successful techniques for semantic segmentation are based on fully convolution networks (FCN)~\cite{long2014fully}. These are adapted from networks designed to classify whole images~\cite{krizhevsky2012imagenet,szegedy2014going,simonyan2014very}, and have demonstrated impressive level of performance. The FCN approach can be thought of as sliding an classification network around an input image, and processes each sliding window area independently. In particular, FCN disregards global information about an image, thus ignoring potentially useful scene-level semantic context. In order to integrate more context, several approaches~\cite{chen2014semantic, schwing2015fully, lin2015efficient, zheng2015conditional}, propose using techniques from graphical models such as conditional random field (CRF), to introduce global context and structured information into a FCN. Although powerful, these architectures can be complex, combining both the challenges of tuning a deep neural network and a CRF, and require a fair amount of experience in managing the idiosyncrasies of training methodology and parameters. At the least, this leads to time-consuming training and inference.

\begin{figure}[tbp]
\centering
\begin{subfigure}[b]{0.109\linewidth}
	\includegraphics[width=\linewidth]{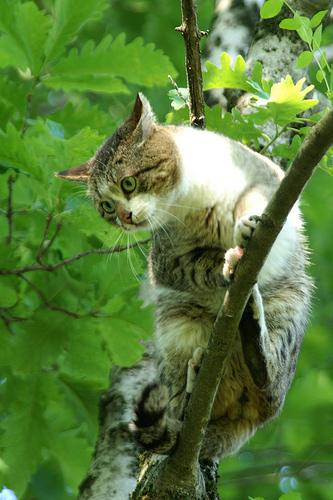}
	\caption{\footnotesize{Image}}
\end{subfigure}
\begin{subfigure}[b]{0.112\linewidth}
	\includegraphics[width=\linewidth]{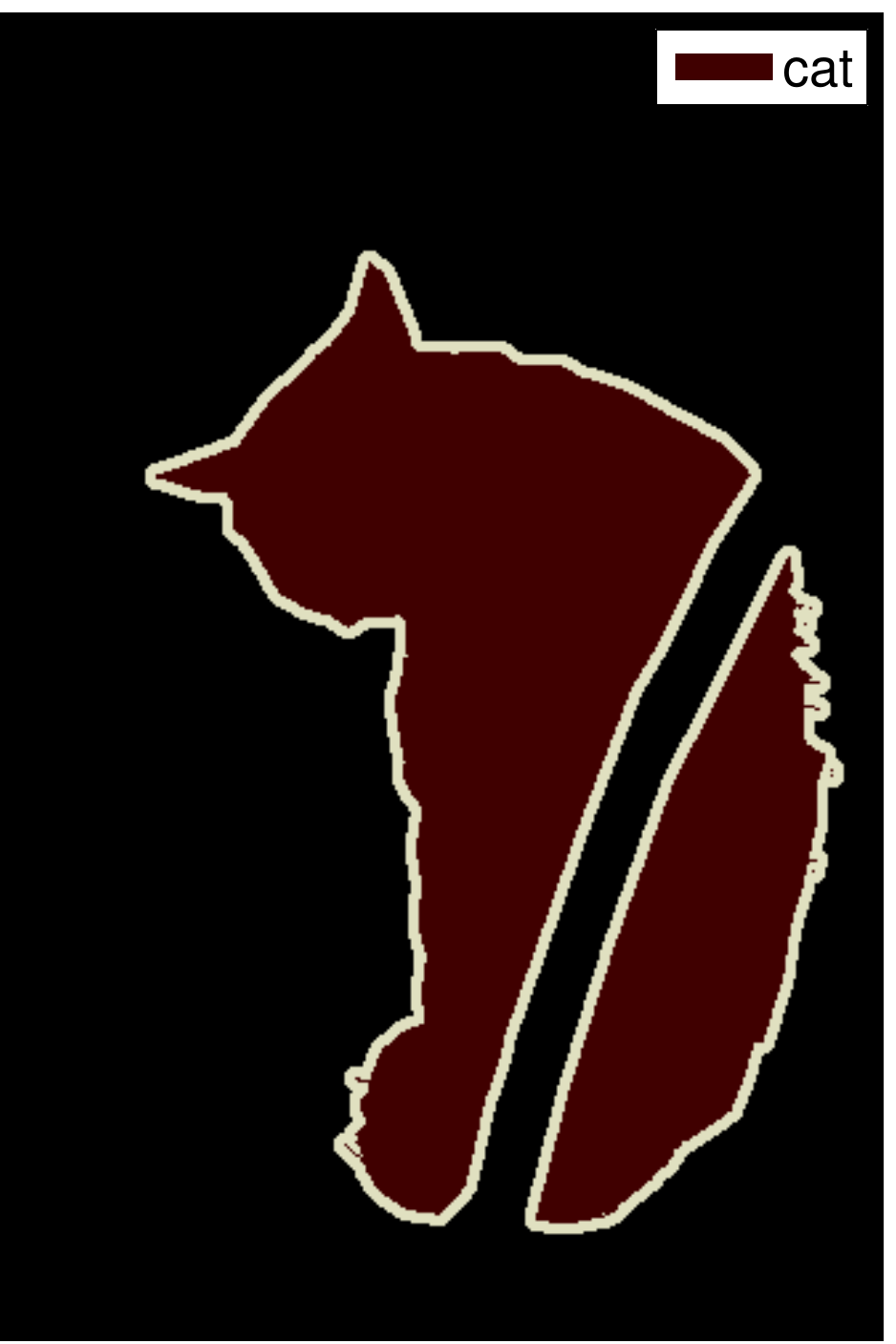}
	\caption{\footnotesize{Truth}}
\end{subfigure}
\begin{subfigure}[b]{0.112\linewidth}
	\includegraphics[width=\linewidth]{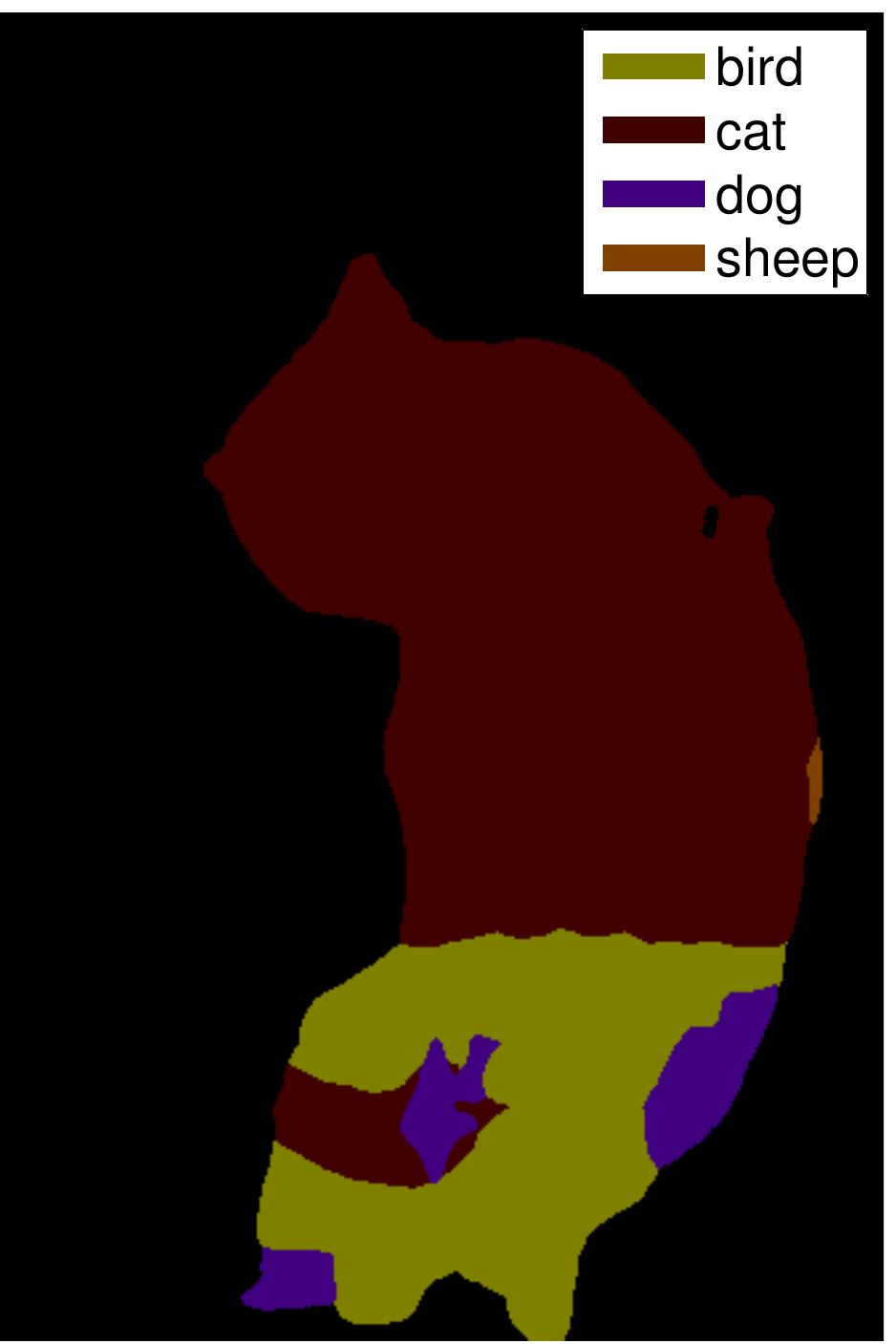}
	\caption{\footnotesize{FCN}}
	\label{fig:fcnoutput}
\end{subfigure}
\begin{subfigure}[b]{0.113\linewidth}
	\includegraphics[width=\linewidth]{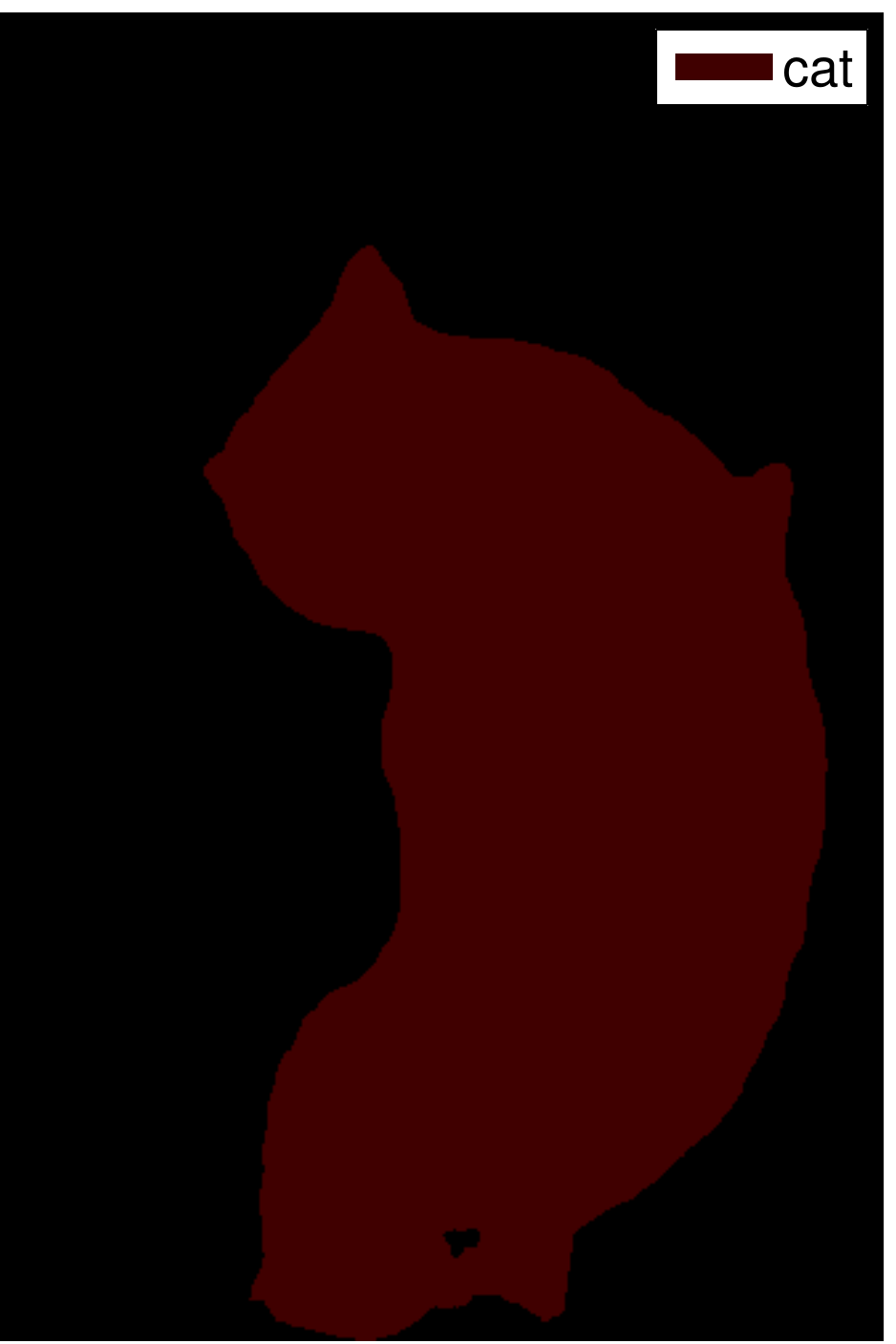}
	\caption{ParseNet}
	\label{fig:parsenetoutput}
\end{subfigure}
\hspace{0.2em}
\begin{subfigure}[b]{0.52\linewidth}
	\includegraphics[width=\linewidth]{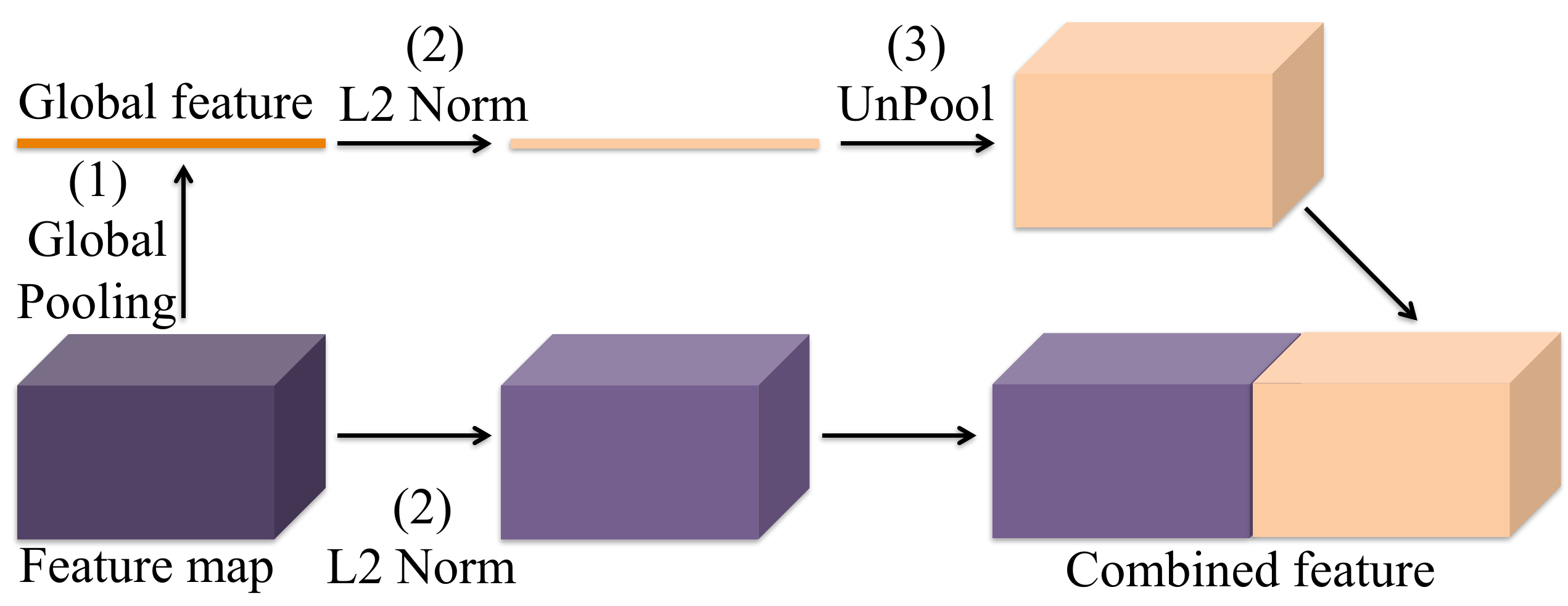}
	\caption{ParseNet contexture module overview.}
\end{subfigure}
\caption{ParseNet uses extra global context to clarify local confusion and smooth segmentation.}
\label{fig:system}
\end{figure}

In this work, we propose \emph{ParseNet}, an end-to-end simple and effective convolutional neural network, for semantic segmentation. One of our main contributions, as shown in Fig.~\ref{fig:system}, is to use global context to help clarify local confusions. Looking back at previous work, adding global context for semantic segmentation is not a new idea, but has so far been pursued in patch-based frameworks~\cite{lucchi2011spatial}. Such patch-based approaches have much in common with detection and segmentation work that have also shown benefits from integrating global context into classifying regions or objects in an image~\cite{szegedy2014scalable, mostajabi2014feedforward}. Our approach allows integrating global context in an end-to-end fully convolutional network (as opposed to a patch-based approach) for semantic segmentation with small computational overhead. In our setting, the image is not divided into regions or objects, instead the network makes a joint prediction of all pixel values. Previous work on fully convolutional networks did not include global features, and there were limits in the pixel distance across which consistency in labeling was maintained.

The key "widget" that allows adding global context to the FCN framework is simple, but has several important consequences in addition to improving the accuracy of FCN. First, the entire end-to-end process is a single deep network, making training relatively straightforward compared to combining deep networks and CRFs. In addition, the way we add global context does not introduce much computational overhead versus training and evaluating a standard FCN, while improving performance significantly. In our approach, the feature map for a layer is pooled over the whole image to result in a context vector. This is appended to each of the features sent on to the subsequent layer of the network. In implementation, this is accomplished by unpooling the context vector and appending the resulting feature map with the standard feature map. The process is shown in Fig.~\ref{fig:system}. This technique can be applied selectively to feature maps within a network, and can be used to combine information from multiple feature maps, as desired. Notice that the scale of features from different layers may be quite different, making it difficult to directly combine them for prediction. We find that $L_2$ normalizing features for each layer and combining them using a scaling factor learned through backpropagation works well to address this potential difficulty. 

In section~\ref{sec:experiment}, we demonstrate that these operations, appending global context pooled from a feature map along with an appropriate scaling, are sufficient to significantly improve performance over the basic FCN, resulting in accuracy on par with the method of~\cite{chen2014semantic} that uses detailed structure information for post processing. That said, we do not advocate ignoring the structure information. Instead, we posit that adding the global feature is a simple and robust method to improve FCN performance by considering contextual information. In fact, our network can be combined with explicit structure output prediction, e.g. a CRF, to potentially further increase performance.

The rest of the paper is organized as follows. In Section~\ref{sec:relatedwork} we review the related work. Our proposed approach is described in Section~\ref{sec:methods} followed by extensive experimental validation in Section~\ref{sec:experiment}. We conclude our work and describe future directions in Section~\ref{sec:futurework}.

\section{Related Work}
\label{sec:relatedwork}
Deep convolutional neural networks (CNN)~\cite{krizhevsky2012imagenet, szegedy2014going, simonyan2014very} have become powerful tools not only for whole image classification, but also for object detection and semantic segmentation~\cite{girshick2014rich, szegedy2014scalable, gupta2014learning}. This success has been attributed to both the large capacity and effective training of the CNN. Following the \textit{proposal + post-classification} scheme~\cite{uijlings2013selective}, CNNs achieve state-of-the-art results on object detection and segmentation tasks. As a caveat, even though a single pass through the networks used in these systems is approaching or already past video frame rate for individual patch, these approaches require classifying hundreds or thousands of patches per image, and thus are still slow. \cite{he2014spatial, long2014fully} improve the computation by applying convolution to the whole image once, and then pool features from the final feature map of the network for each region proposal or pixel to achieve comparable or even better results. Yet, these methods still fall short of including whole image context and only classify patches or pixels locally. Our ParseNet is built upon the fully convolutional network architecture~\cite{long2014fully} with a strong emphasis on including contextual information in a simple approach.

For semantic segmentation, using context information~\cite{rabinovich2007objects, shotton2009textonboost, torralba2003contextual} from the whole image can significantly help classifying local patches. \cite{lucchi2011spatial} shows that by concatenating features from the whole image to the local patch, the inclusion of post processing (i.e. CRF smoothing) becomes unnecessary because the image level features already encode the smoothness. \cite{mostajabi2014feedforward} demonstrate that by using the "zoom-out" features, which is a combination of features for each super pixel, region surrounding it, and the whole image, they can achieve impressive performance for the semantic segmentation task. These approaches pool features differently for local patches and the whole image, making it difficult to train the whole system end-to-end. Exploiting the FCN architecture, ParsetNet can directly use global average pooling from the final (or any) feature map, resulting in the feature of the whole image, and use it as context. Experiments results confirm that ParseNet can capture the context of the image and thus improve local patch prediction results.

There is another line of work that attempts to combine graphical models with CNNs to incorporate both context and smoothness priors. \cite{chen2014semantic} first uses a FCN to estimate the unary potential, then applies a fully connected CRF to smooth the predictions spatially. As this approach consists of two decoupled stages, it is difficult to train the FCN properly to minimize the final objective of smooth and accurate semantic segments. A more unified and principled approach is to incorporate the structure information during training directly. ~\cite{schwing2015fully} propagates the marginals computed from the structured loss to update the network parameters, ~\cite{lin2015efficient} uses piece-wise training to make learning more efficient by adding a few extra piece-wise networks, while~\cite{zheng2015conditional} convert CRF learning to recurrent neural network (RNN) and use message passing to do the learning and inference. However, we show that our method can achieve comparable accuracy, with a simpler -- hence more robust -- structure, while requiring only a small amount of additional training/inference time.
\begin{figure}
	\centering
	\includegraphics[width=0.2\linewidth]{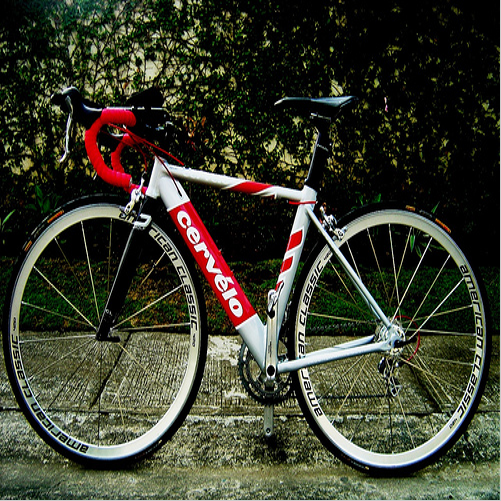}\hspace{2em}
	\includegraphics[width=0.2\linewidth]{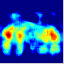}\hspace{2em}
	\includegraphics[width=0.2\linewidth]{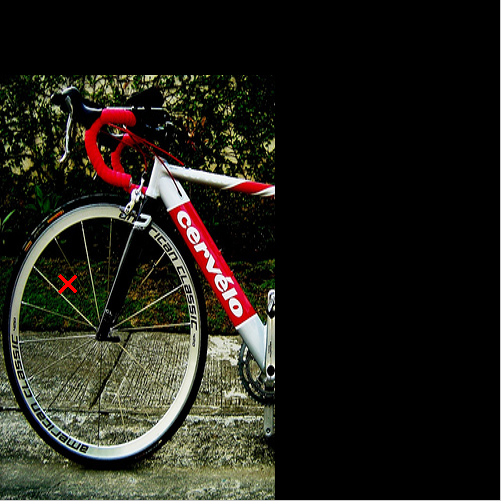}\hspace{2em}
	\includegraphics[width=0.2\linewidth]{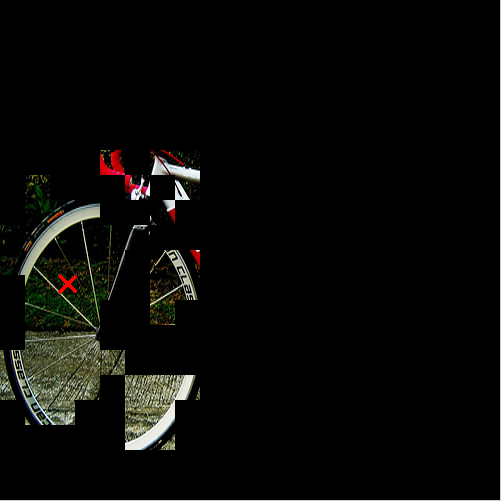}\\
	\begin{subtable}[b]{0.2\linewidth}
		\includegraphics[width=\linewidth]{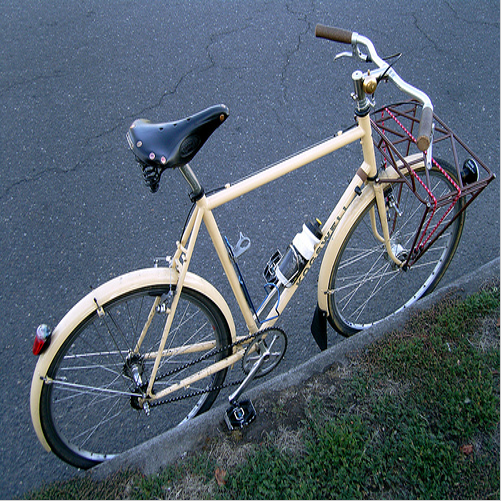}
		\caption{Original Image}
	\end{subtable}
    \hspace{1.8em}
	\begin{subtable}[b]{0.2\linewidth}
		\includegraphics[width=\linewidth]{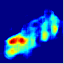}
		\caption{Activation map}
	\end{subtable}
    \hspace{1.8em}
	\begin{subtable}[b]{0.2\linewidth}
		\includegraphics[width=\linewidth]{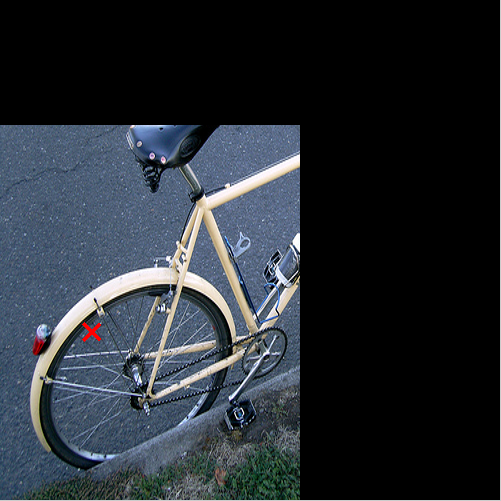}
		\caption{Theoretical RF}
	\end{subtable}
    \hspace{1.8em}
	\begin{subtable}[b]{0.2\linewidth}
		\includegraphics[width=\linewidth]{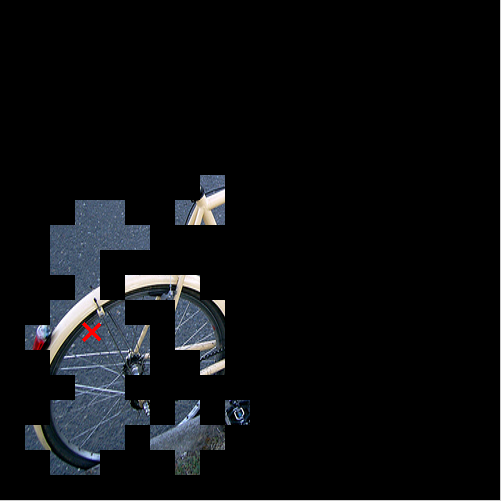}
		\caption{Empirical RF}
	\end{subtable}
	\caption{\textbf{Receptive field (RF) size for last layer.} (a) original image; (b) activation map on bicycle from a channel of the last layer of a network; (c) theoretical receptive field of the maximum activation (marked by red cross) is defined by the network structure; (d) empirical receptive field affecting the activation. Clearly empirical receptive field is not large enough to capture the global context.}
\label{fig:receptivefield}
\end{figure}
\section{ParseNet}
\label{sec:methods}
\subsection{Global Context}
Context is known to be very useful for improving performance on detection and segmentation tasks using deep learning. \cite{mostajabi2014feedforward, szegedy2014scalable} and references therein illustrate how context can be used to help in different tasks. As for semantic segmentation, per pixel classification, is often ambiguous in the presence of only local information. However, the task becomes much simpler if contextual information, from the whole image, is available. Although theoretically, features from the top layers of a network have very large receptive fields (e.g. fc7 in FCN with VGG has a $404 \times 404$ pixels receptive field), we argue that in practice, the empirical size of the receptive fields is much smaller, and is not enough to capture the global context. To identify the effective receptive field, we slide a small patch of random noise across the input image, and measure the change in the activation of the desired layer. If the activation does not vary significantly, that suggests the given random patch is outside of the empirical receptive field, as shown in Figure~\ref{fig:receptivefield}. The effective receptive field at the last layer of this network barely covers $\frac{1}{4}$ of the entire image. Such an effect of difference between empirical and theoretical receptive field sizes was also observed in~\cite{zhou2014object}. Fortunately, it is rather straightforward to get the context within the FCN architecture. Specifically, we use global average pooling and pool the context features from the last layer or any layer if that is desired. The quality of semantic segmentation is greatly improved by adding the global feature to local feature map, either with early fusion~\footnote{we use unpool operation by simply replicating the global feature horizontally and vertically to have the same size as the local feature map.} or late fusion as discussed in Sec.~\ref{sec:earlylatefusion}. For example, Fig~\ref{fig:system} has misclassified a large portion of the image as bird since it only used local information, however, adding contextual information in the loop, which might contain strong signal of cat, corrects the mistake. Experiment results on VOC2012 and PASCAL-Context dataset also verify our assumption. Compared with~\cite{chen2014semantic}, the improvement is similar as of using CRF to post-process the output of FCN.

In addition, we also tried to follow the spatial pyramid idea~\cite{lazebnik2006beyond} to pool features from increasingly finer sub-regions and attach them to local features in the sub-regions, however, we did not observe significant improvements. We conjecture that it is because the (empirical) receptive field of high-level feature maps is larger than or similar as those sub-regions. However features pooled from the whole image are still beneficial.

\subsection{Early Fusion and Late Fusion}
\label{sec:earlylatefusion}
Once we get the global context feature, there are two general standard paradigms of using it with the local feature map. First, the \emph{early fusion}, illustrated in in Fig.~\ref{fig:system} where we unpool (replicate) global feature to the same size as of local feature map spatially and then concatenate them, and use the combined feature to learn the classifier. The alternative approach, is \emph{late fusion}, where each feature is used to learn its own classifier, followed by merging the two predictions into a single classification score~\cite{long2014fully, chen2014semantic}. There are cons and pros for both fusion methods. If there is no additional processing on combined features, \emph{early fusion} is quite similar to \emph{late fusion} as pointed out in~\cite{hariharan2014hypercolumns}. With \emph{late fusion}, there might be a case where individual features cannot recognize something but combining them may and there is no way to recover from independent predictions. Our experiments show that both method works more or less the same if we normalize the feature properly for early fusion case.

When merging the features, one must be careful to normalize each individual feature to make the combined feature work well; in classical computer vision this is referred as the cue combination problem. As shown in Fig.~\ref{fig:differentscale}, we extract a feature vector at a position combined from increasing higher level layers (from left to right), with lower level feature having a significantly larger scale than higher level layers. As we show in Sec.~\ref{sec:normalize}, by naively combining features, the resultant feature will not be discriminative, and heavy parameter tuning will be required to achieve sufficient accuracy. Instead, we can first $L_2$ normalize each feature and also possibly learn the scale parameter, which makes the learning more stable. We will describe more details in Sec.~\ref{sec:l2norm}.

\subsection{$L_2$ Normalization Layer}
\label{sec:l2norm}
\begin{figure}
	\centering
	\includegraphics[width=0.4\linewidth]{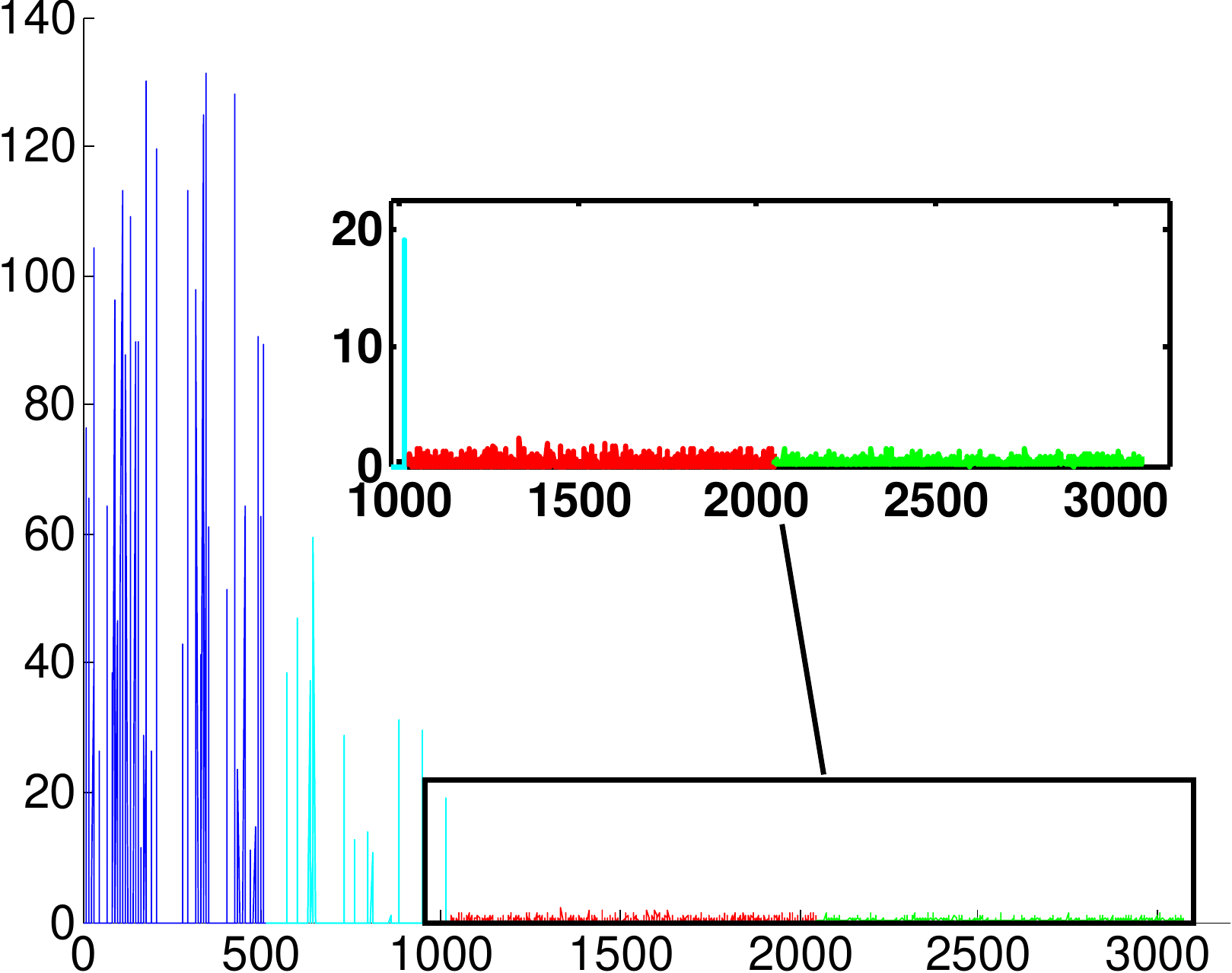}
	\caption{Features from 4 different layers have activations that are of drastically different scales. Each color corresponds to a different layers' feature. While \emph{\textcolor[rgb]{0,0,1}{blue}} and \emph{\textcolor[rgb]{0,1,1}{cyan}} are on a comparable scale, \emph{\textcolor[rgb]{1,0,0}{red}} and \emph{\textcolor[rgb]{0,1,0}{green}} features are of a scale 2 orders of magnitude less.}
	\label{fig:differentscale}
\end{figure}

As discussed above and shown in Fig.~\ref{fig:differentscale}, we need to combine two (or more) feature vectors, which generally have different scale and norm. Naively concatenating features leads to poor performance as the "larger" features dominate the "smaller" ones. Although during training, the weight might adjust accordingly, it requires very careful tuning of parameters and depends on dataset, thus goes against the robust principle. We find that by normalizing each individual feature first, and also learn to scale each differently, it makes the training more stable and improves performance.

$L_2$ norm layer is not only useful for feature combination. As was pointed out above, in some cases \emph{late fusion} also works equally well, but only with the help of $L_2$ normalization. For example, if we want to use lower level feature to learn classifier, as demonstrated in Fig.~\ref{fig:differentscale}, some of the features will have very large norm. It is not trivial to learn with it without careful weight initialization and parameter tuning. A work around strategy is to apply an additional convolutional layer~\cite{chen2014semantic, hariharan2014hypercolumns} and use several stages of finetuning~\cite{long2014fully} with much lower learning rate for lower layer. This again goes against the principle of simply and robustness. In our work, we apply $L_2$-norm and learn the scale parameter for each channel before using the feature for classification, which leads to more stable training.

Formally, let $\ell$ be the loss we want to minimize. Here we use the summed softmax loss. For a layer with $d$-dimensional input $\mathbf{x} = (x_1\cdots x_d)$, we will normalize it using $L_2$-norm\footnote{We have only tried $L_2$ norm, but can also potentially try other $l_p$ norms.} with $\hat{\mathbf{x}} = \frac{\mathbf{x}}{||\mathbf{x}||_2}$ where $||\mathbf{x}||_2 = \Big(\sum_{i=1}^d |x_i|^2\Big)^{1/2}$ is the $L_2$ norm of $\mathbf{x}$.

Note that simply normalizing each input of a layer changes the scale of the layer and will slow down the learning if we do not scale it accordingly. For example, we tried to normalize a feature s.t. $L_2$-norm is 1, yet we can hardly train the network because the features become very small. However, if we normalize it to e.g. $10$ or $20$, the network begins to learn well. Motivated by batch normalization~\cite{ioffe2015batch} and PReLU~\cite{he2015delving}, we introduce a scaling parameter $\gamma_i$, for each channel, which scales the normalized value by $y_i = \gamma_i \hat{x}_i$.

The number of extra parameters is equal to total number of channels, and are negligible and can be learned with backprogation. Indeed, by setting $\gamma_i = ||\mathbf{x}||_2$, we could recover the $L_2$ normalized feature, if that was optimal. Notice that this is simple to implement as the normalization and scale parameter learning only depend on each input feature vector and do not need to aggregate information from other samples as batch normalization does. During training, we use backpropagation and chain rule to compute derivatives with respect to scaling factor $\mathbf{\gamma}$ and input data $\mathbf{x}$
\begin{equation}
	\frac{\partial \ell}{\partial \hat{\mathbf{x}}} = \frac{\partial \ell}{\partial \mathbf{y}} \cdot \mathbf{\gamma}\qquad\qquad
	\frac{\partial \ell}{\partial \mathbf{x}} = \frac{\partial \ell}{\partial \hat{\mathbf{x}}} \Big(\frac{\mathbf{I}}{||\mathbf{x}||_2} - \frac{\mathbf{x}\mathbf{x}^T}{||\mathbf{x}||_2^3}\Big)\qquad\qquad
	\frac{\partial \ell}{\partial \gamma_i} = \sum_{y_i} \frac{\partial \ell}{\partial y_i} \hat{x}_i
\end{equation}
For our case, we need to do $L_2$-norm per each pixel in a feature map instead of the whole. We can easily extend the equations by doing it elemental wise as it is efficient.

\section{Experiments}
\label{sec:experiment}
In this section, we mainly report results on three benchmark datasets: VOC2012~\cite{everingham2014pascal} and PASCAL-Context~\cite{mottaghi2014role}. VOC2012 has 20 object classes and one background class. Following~\cite{long2014fully, chen2014semantic}, we augment it with extra annotations from Hariharan \etal~\cite{hariharan2011semantic} that leads to 10,582, 1,449, and 1,456 images for training, validation, and testing. PASCAL-Context~\cite{mottaghi2014role} fully labeled all scene classes appeared in VOC2010. We follow the same training + validation split as defined and used in~\cite{mottaghi2014role, long2014fully}, resulting in 59 object + stuff classes and one background classes with 4,998 and 5105 training and validation images. All the results we describe below use the training images to train, and most of the results are on the validation set. We also report results on VOC2012 test set. We use Caffe~\cite{jia2013caffe} and fine-tune ParseNet from VGG-16 network~\cite{simonyan2014very} for different dataset.

\subsection{Best practice of finetuning}
As we know parameters are important for training/finetuning network, we try to reproduce the state-of-the-art systems' results by exploring the parameter space and achieve better baseline performance.

\noindent\textbf{PASCAL-Context} We start from the public system FCN-32s PASCAL-Context. Notice that it uses the accumulated gradient and affine transformation tricks that were introduced in~\cite{long2014fully}. As such, it can deal with any input image of various sizes without warping or cropping it to fixed size, which can distort the image and affect the final segmentation result. Table~\ref{tab:pascalcontextbase} shows our different versions of reproduced baseline results. Baseline A uses the exactly same protocol, and our result is 1.5\% lower. In Baseline B, we tried more iteration (160k vs. 80k) of finetuning and achieved similar performance to the reported one. Then, we modified the network a bit, i.e. we used "xavier" initialization~\cite{glorot2010understanding}, higher base learning rate (1e-9 vs. 1e-10), and lower momentum (0.9 vs. 0.99), and we achieved $1\%$ higher accuracy as shown in Baseline C. What's more, we also remove the 100 padding in the first convolution layer and observed no significant difference but network trained slightly faster. Furthermore, we also used "poly" learning rate policy ($\text{base\_lr}\times (1 - \frac{\text{iter}}{\text{max\_iter}})^{\text{power}}$, where power is set to 0.9.) as it is proved to converge faster than normal "step" policy, and thus can achieve $1.5\%$ better performance with the same iterations (80k). All experimental results on PASCAL-Context are shown in table~\ref{tab:pascalcontextbase}.

\begin{table}[!htb]
	\centering
	\begin{tabular}{c|c}
		PASCAL-Context & Mean IoU\\
		\hline\hline
		FCN-32s~\tablefootnote{\url{https://gist.github.com/shelhamer/80667189b218ad570e82\#file-readme-md}} & 35.1\\
		\hline
		Baseline A & 33.57\\
		Baseline B & 35.04\\
		Baseline C & 36.16\\
		Baseline D & \textbf{36.64}\\
		\hline
	\end{tabular}
	\caption{\textbf{Reproduce FCN-32s on PASCAL-Context.} There are various modifications of the architecture that are described in Section $4.1$.}\label{tab:pascalcontextbase}
\end{table}

\noindent\textbf{PASCAL VOC2012} We carry over the parameters we found on PASCAL-Context to VOC2012. We tried both FCN-32s and DeepLab-LargeFOV\footnote{\url{https://bitbucket.org/deeplab/deeplab-public/}}. Table~\ref{tab:voc2012base} shows the reproduced baseline results. DeepLab is very similar to FCN-32s, and our reproduced result is 5\% better (64.96 vs. 59.80) using the parameters we found in PASCAL-Context. DeepLab-LargeFOV uses the filter rarefication technique (atrous algorithm) that has much less parameters and is faster. We also use the same parameters on this architecture and can achieve 3.5\% improvements. The gap between these two models is not significant anymore as reported in~\cite{chen2014semantic}. Later on, we renamed DeepLab-LargeFOV Baseline as ParseNet Baseline, and ParseNet is ParseNet Baseline plus global context.

\begin{table}[!htb]
	\centering
	\begin{tabular}{c|c}
		VOC2012 & Mean IoU\\
		\hline\hline
		DeepLab~\cite{chen2014semantic} & 59.80\\
		DeepLab-LargeFOV~\cite{chen2014semantic} & 62.25\\
		\hline
		DeepLab Baseline & 64.96\\
		DeepLab-LargeFOV Baseline & \textbf{65.82}\\
		\hline
	\end{tabular}
	\caption{\textbf{Reproduce DeepLab and DeepLab-LargeFOV on PASCAL VOC2012.}}\label{tab:voc2012base}
\end{table}

Until now, we see that parameters and details are important to get best performance using FCN models. Below, we report all our results with the reproduced baseline networks.

\subsection{Combining Local and Global Features}
\label{sec:normalize}
In this section, we report results of combining global and local feature on three dataset: SiftFlow~\cite{liu2011nonparametric}, PASCAL-Context, and PASCAL VOC2012. For simplicity, we use pool6 as the global context feature, conv5 as conv5\_3, conv4 as conv4\_3, and conv3 as conv3\_3 through the rest of paper.

\noindent\textbf{SiftFlow} is a relatively small dataset that only has 2,688 images with 33 semantic categories. We do not use the geometric categories during training. We use the FCN-32s network with the parameters found in PASCAL-Context. Instead of using two stages of learning as done in~\cite{long2014fully}, we combine the feature directly from different layers for learning. As shown in Table~\ref{tab:siftflow}, adding more layers can normally improve the performance as lower level layers have more detailed information. We also notice that adding global context feature does not help much. This is perhaps due to the small image size ($256\times 256$), as we know even the empirical receptive field of fc7 (e.g. Fig.~\ref{fig:receptivefield}) is similar as if not bigger than that, thus pool6 is essentially a noop.
\begin{table}[!htb]
	\centering
	\begin{tabular}{r|p{1.5em}p{1.5em}p{1.5em}p{1.5em}}
		& pixel acc. & mean acc. & mean IU & f.w. IU\\
		\hline\hline
		FCN-16s~\cite{long2014fully} & 85.2 & 51.7 & 39.5 & 76.1\\
		\hline
		fc7 & 85.1 & 44.1 & 35.4 & 75.6\\
		pool6 + fc7 & 85.7 & 43.9 & 35.5 & 76.4\\
		pool6 + fc7 + conv5 & 85.4 & 51.4 & 38.7 & 76.3\\
		pool6 + fc7 + conv5 + conv4 & \textbf{86.8} & \textbf{52.0} & \textbf{40.4} & \textbf{78.1}\\
		\hline
	\end{tabular}
	\caption{\textbf{Results on SiftFlow.} Early fusion can work equally well as late fusion as used in~\cite{long2014fully}. Adding more layers of feature generally increase the performance. Global feature is not that helpful as receptive field size of fc7 is large enough to cover most of the input image.}\label{tab:siftflow}
\end{table}
\vspace{-0.5em}

\noindent\textbf{PASCAL-Context} We then apply the same model on PASCAL-Context by concatenating features from different layers of the network. As shown in Table~\ref{tab:pascalcontext}, by adding global context pool6, it instantly helps improve by about $1.6\%$, which means that context is useful here as opposed to the observation in SiftFlow. Context becomes more important proportionally to the image size. Another interesting observation from the table is that, without normalization, the performance keep increasing until we add conv5. However, if we naively keep adding conv4, it starts decreasing the performance a bit; and if we add conv3, the network collapses. Interestingly, if we normalize all the features before we combine them, we don't see such a drop, instead, adding all the feature together can achieve the state-of-the-art result on PASCAL-Context as far as we know.
\begin{table}[!htb]\small
	\centering
	\begin{tabular}{r|c|c}
		& w/o Norm & w/ Norm\\
		\hline\hline
		FCN-32s & 36.6 & N/A\\
		FCN-8s & 37.8 & N/A\\
		\hline
		fc7 & 36.6 & 36.2\\
		pool6 + fc7 & 38.2 & 37.6\\
		pool6 + fc7 + conv5 & 39.5 & 39.9\\
		pool6 + fc7 + conv5 + conv4 & 36.5 & 40.2\\
		pool6 + fc7 + conv5 + conv4 + conv3 & 0.009 & \textbf{40.4}\\
		\hline
	\end{tabular}
	\caption{\textbf{Results on PASCAL-Context.} Adding more layers helps if we $L_2$ normalize them.}\label{tab:pascalcontext}
\end{table}
\vspace{-0.5em}

\noindent\textbf{PASCAL VOC2012} Since we have reproduced both network architecture on VOC2012, we want to see how does global context, normalization, and early or late fusion affect performance.

We start with using DeepLab Baseline, and try to add pool6 to it. It improves from 64.92\% to 67.49\% by adding pool6 with normalization. Interestingly, without normalizing fc7 and pool6, we don't see any improvements. As opposed to what we observed from SiftFlow and PASCAL-Context. We hypothesize this is due to images in VOC2012 mostly have one or two objects in the image versus the other two dataset who have multiple labels per image, and we need to adjust the weight more carefully to make the context feature more useful.

\begin{figure}
	\centering
	\includegraphics[width=0.24\linewidth]{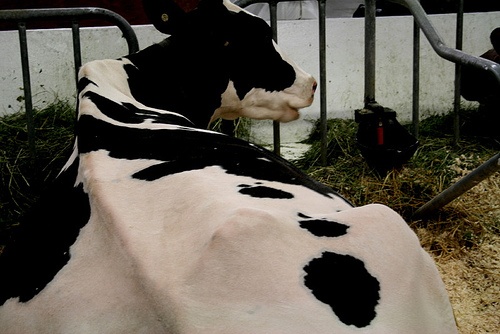}
	\includegraphics[width=0.24\linewidth]{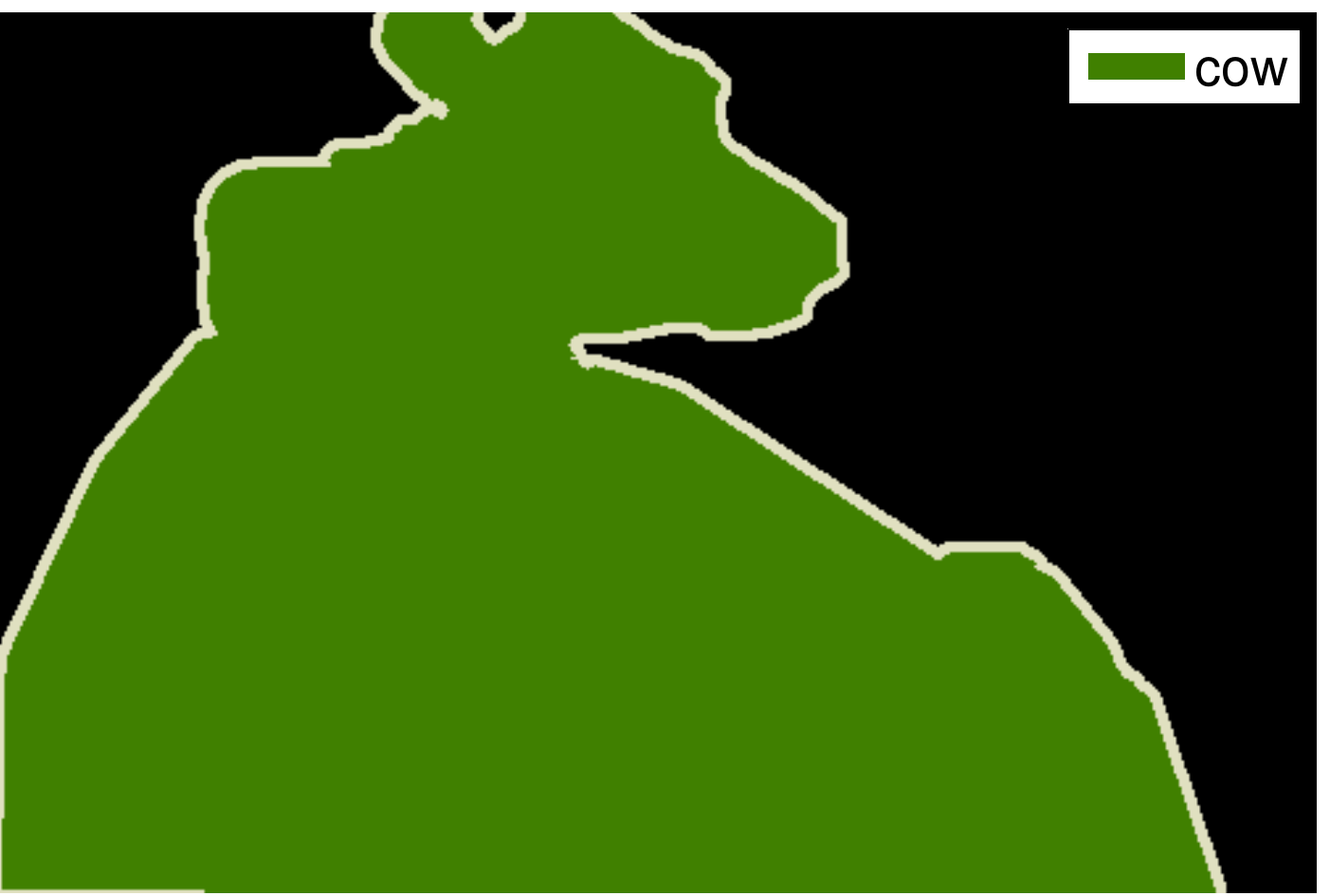}
	\includegraphics[width=0.24\linewidth]{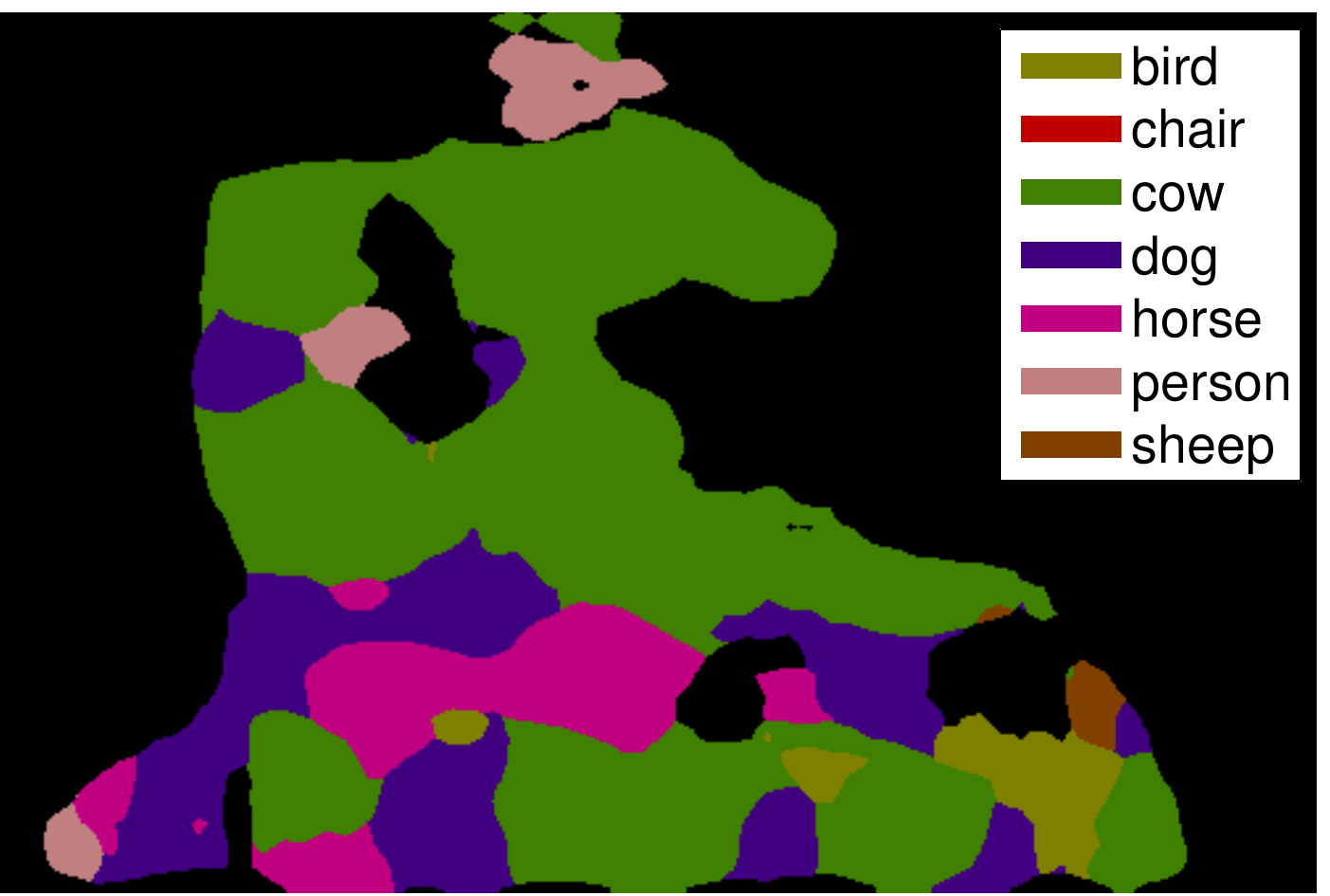}
	\includegraphics[width=0.24\linewidth]{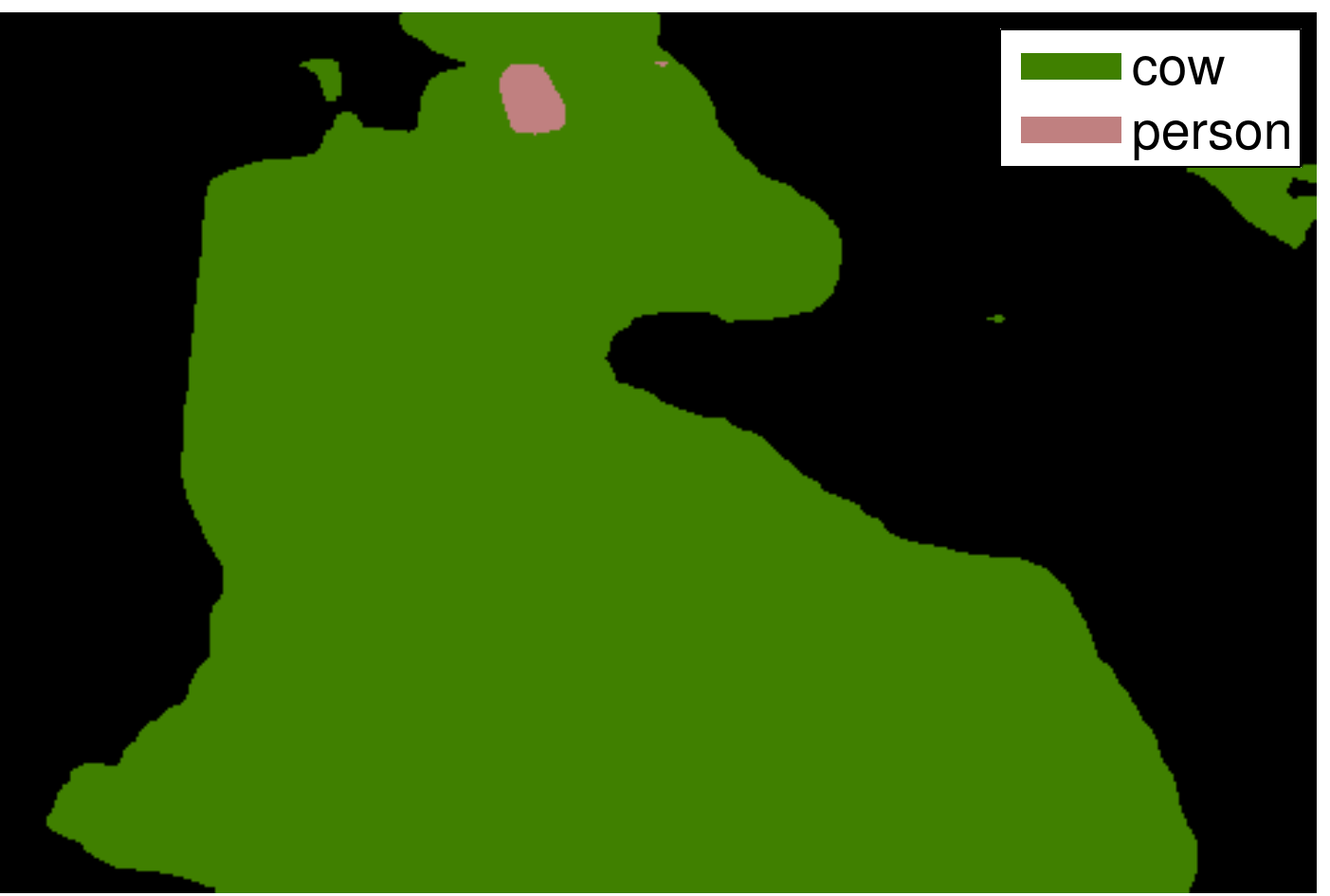}\\
	\includegraphics[width=0.24\linewidth]{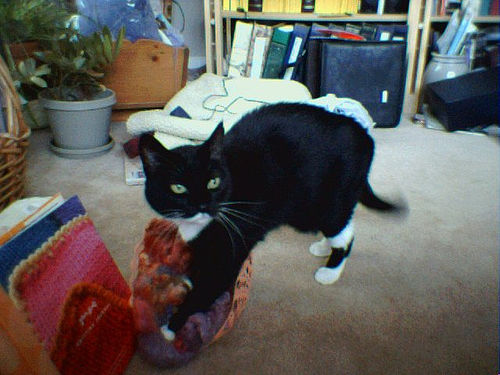}
	\includegraphics[width=0.24\linewidth]{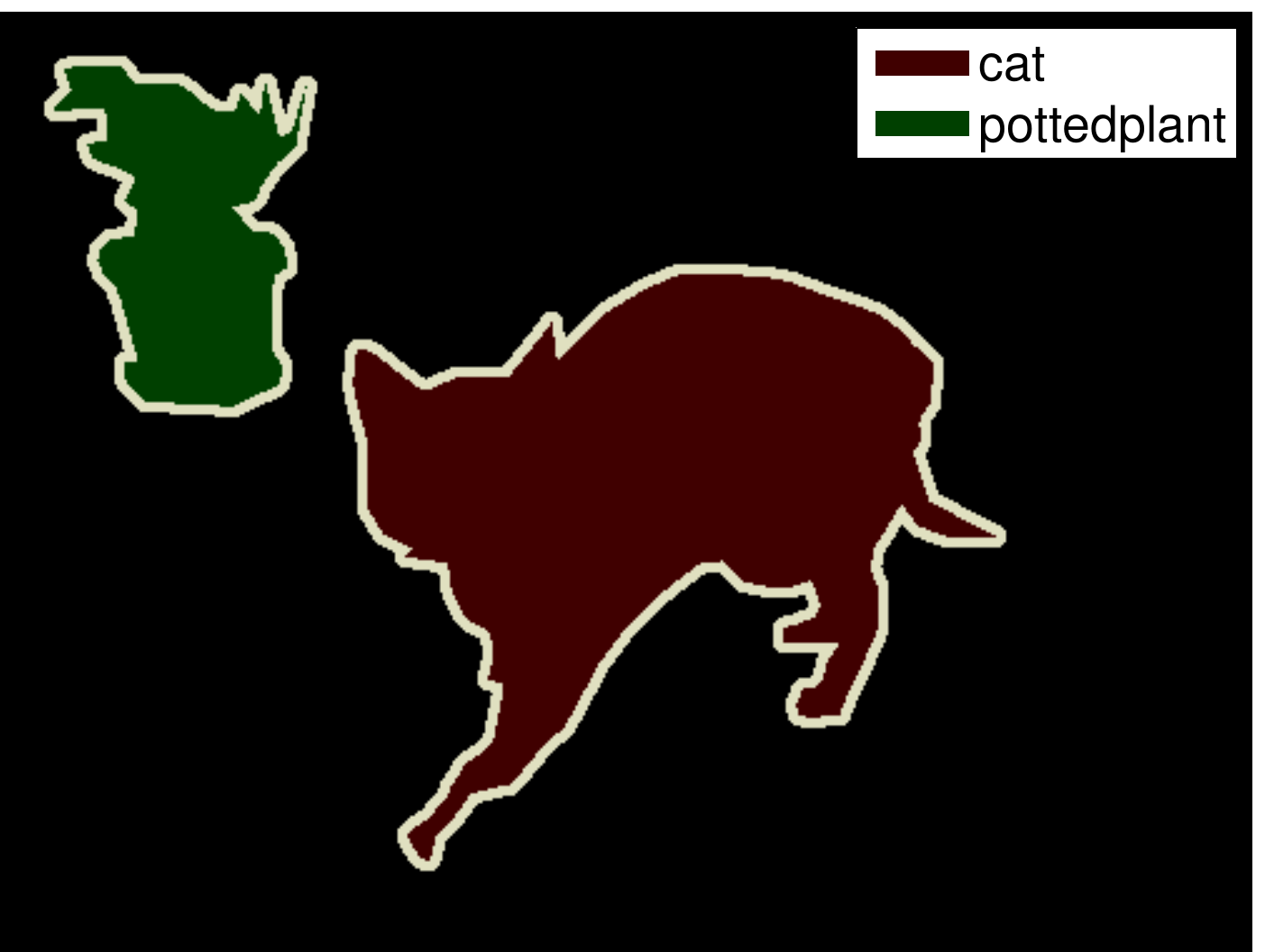}
	\includegraphics[width=0.24\linewidth]{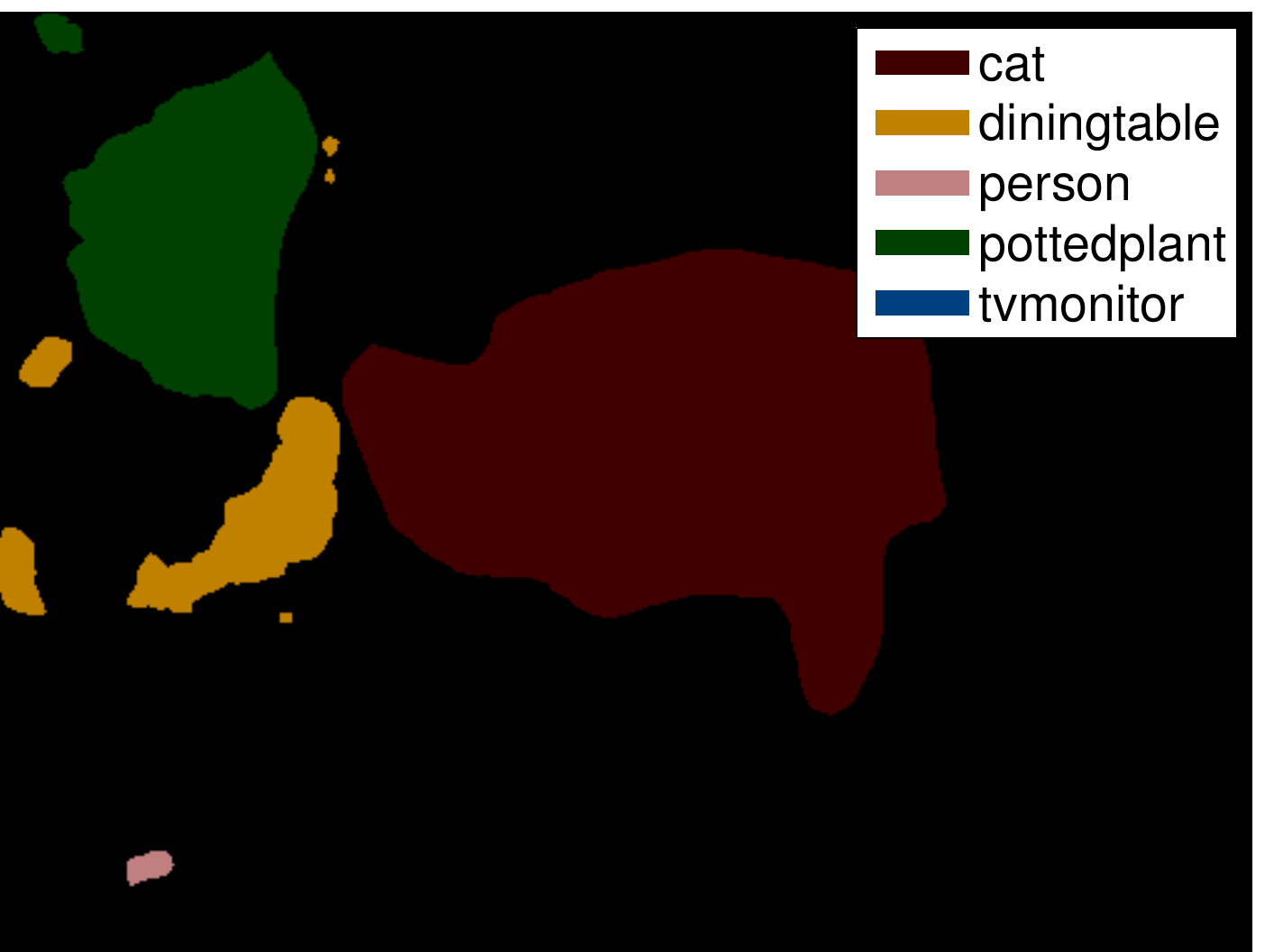}
	\includegraphics[width=0.24\linewidth]{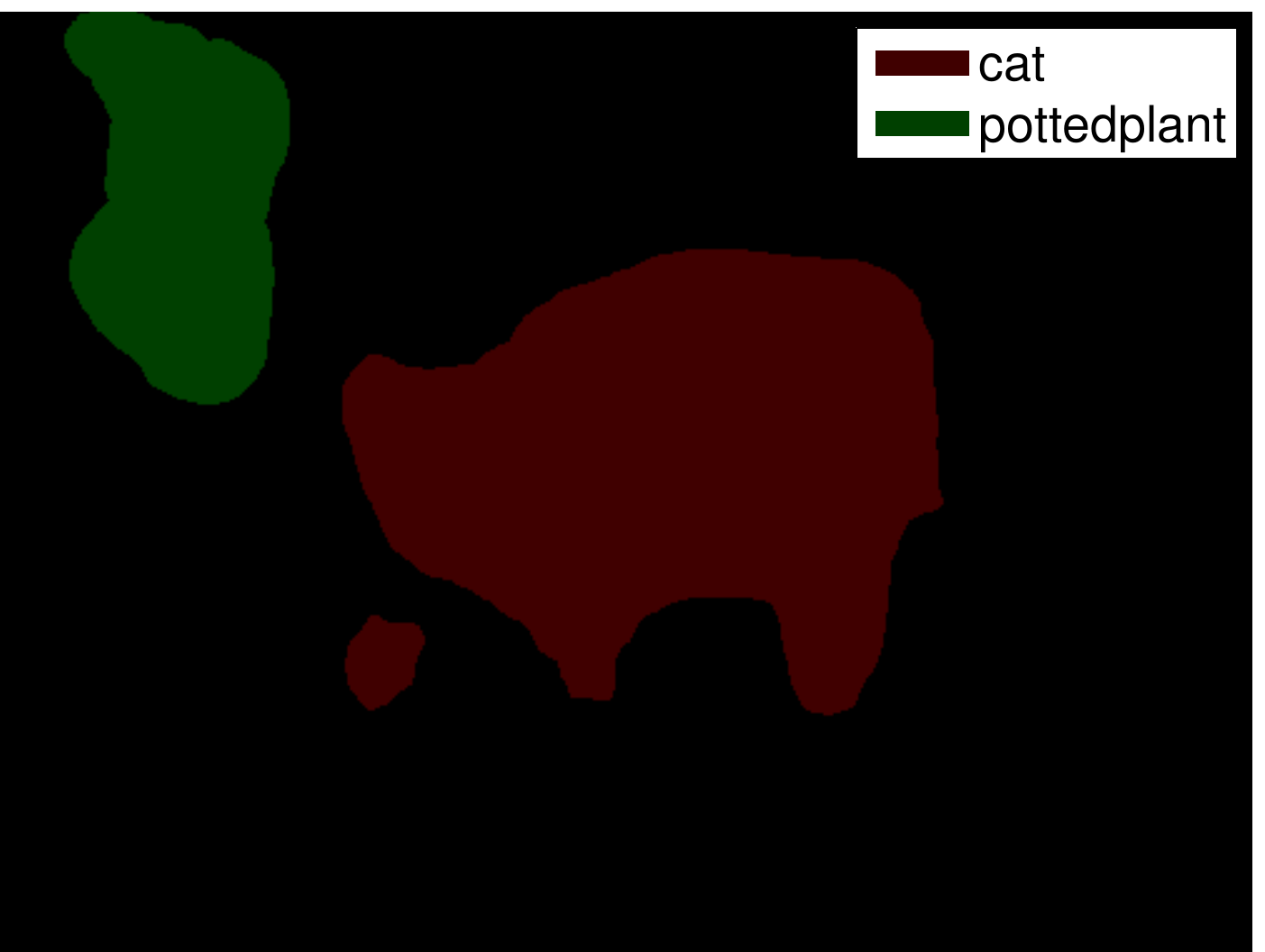}\\
	\includegraphics[width=0.24\linewidth]{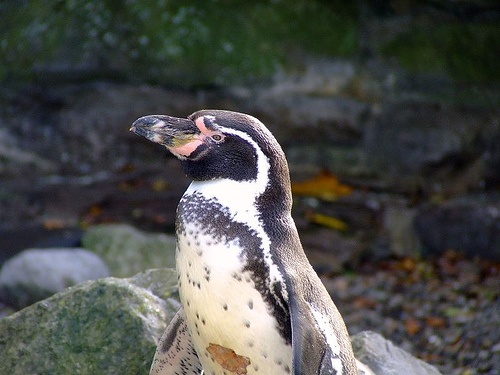}
	\includegraphics[width=0.24\linewidth]{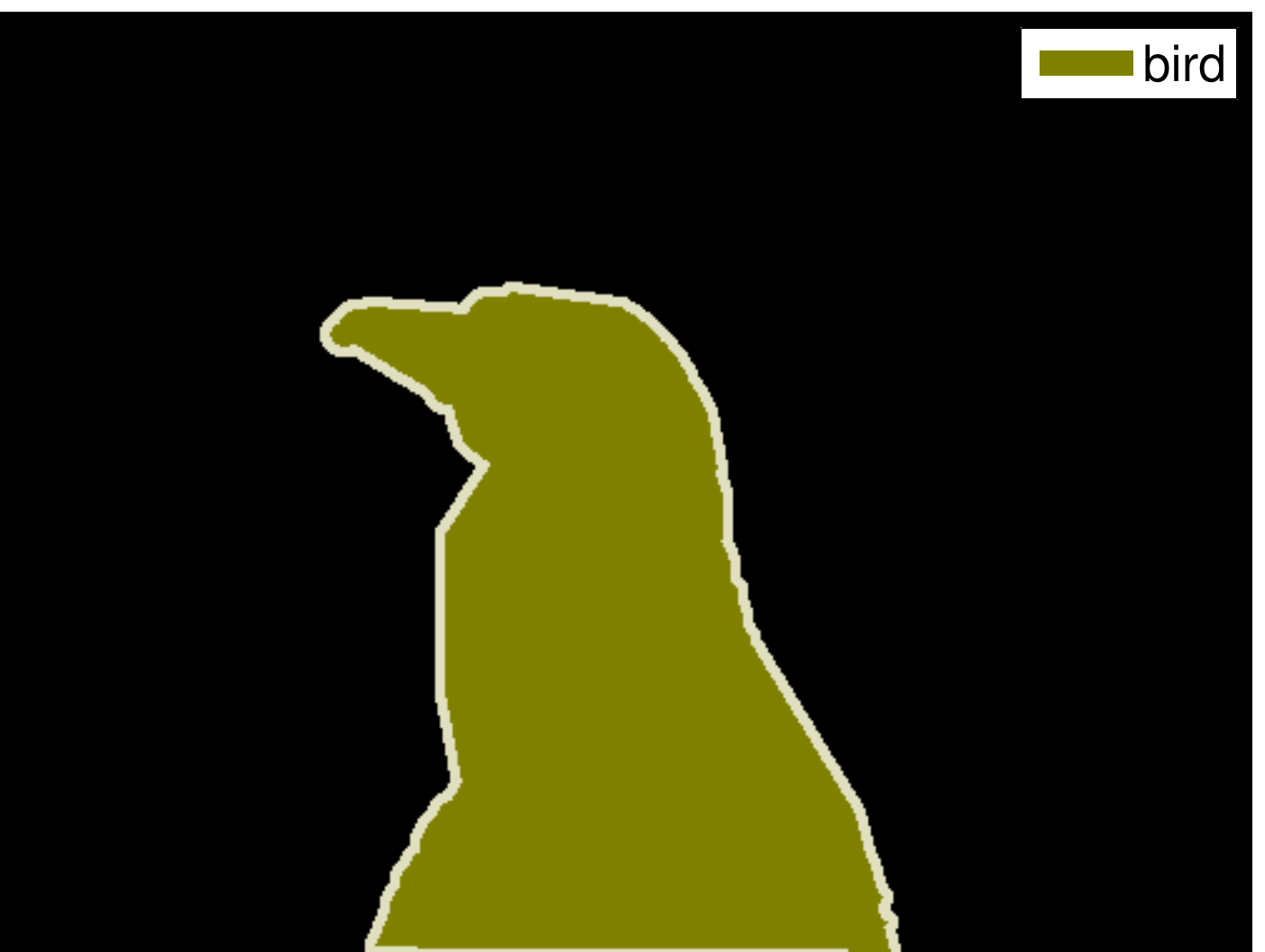}
	\includegraphics[width=0.24\linewidth]{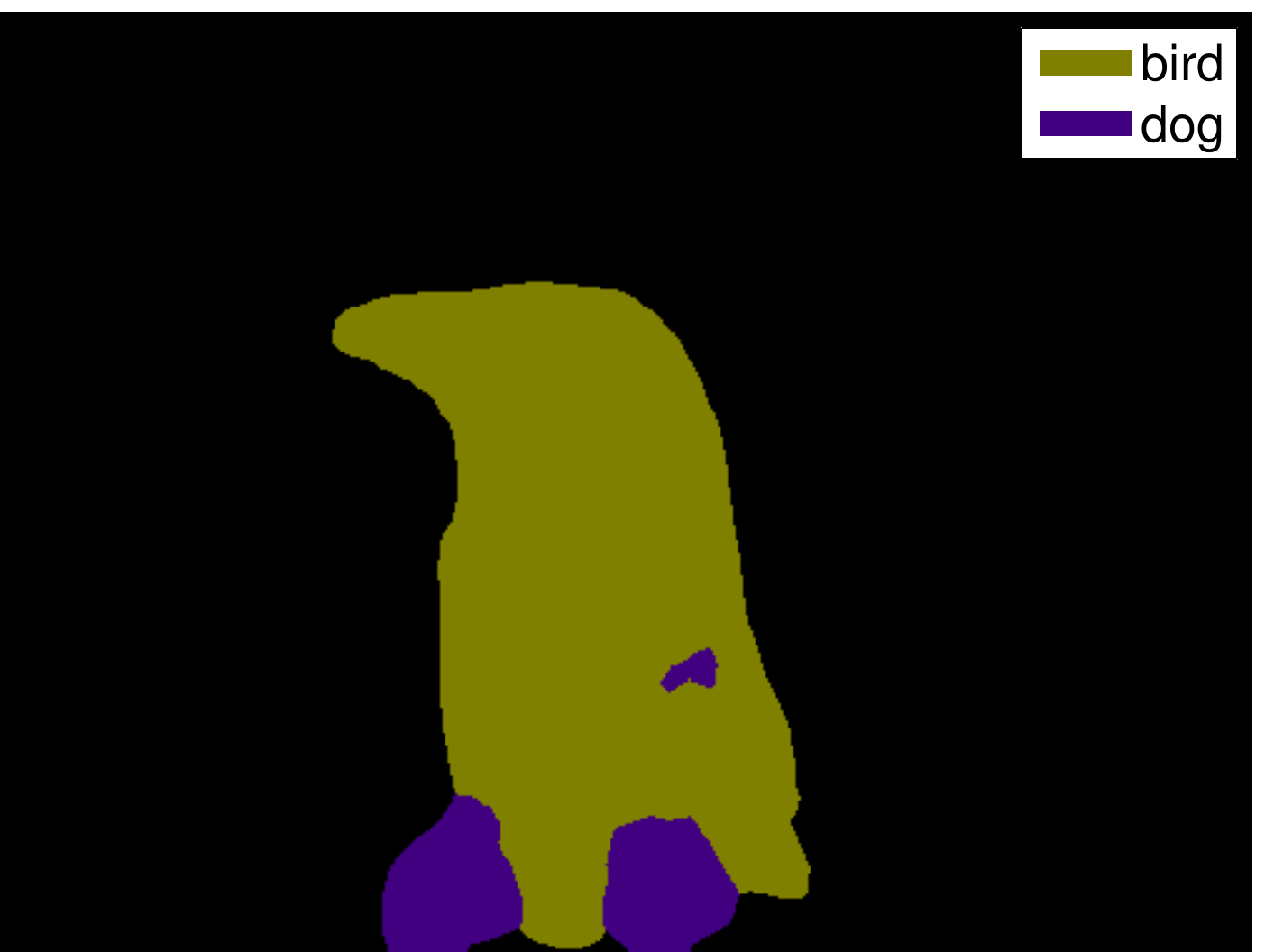}
	\includegraphics[width=0.24\linewidth]{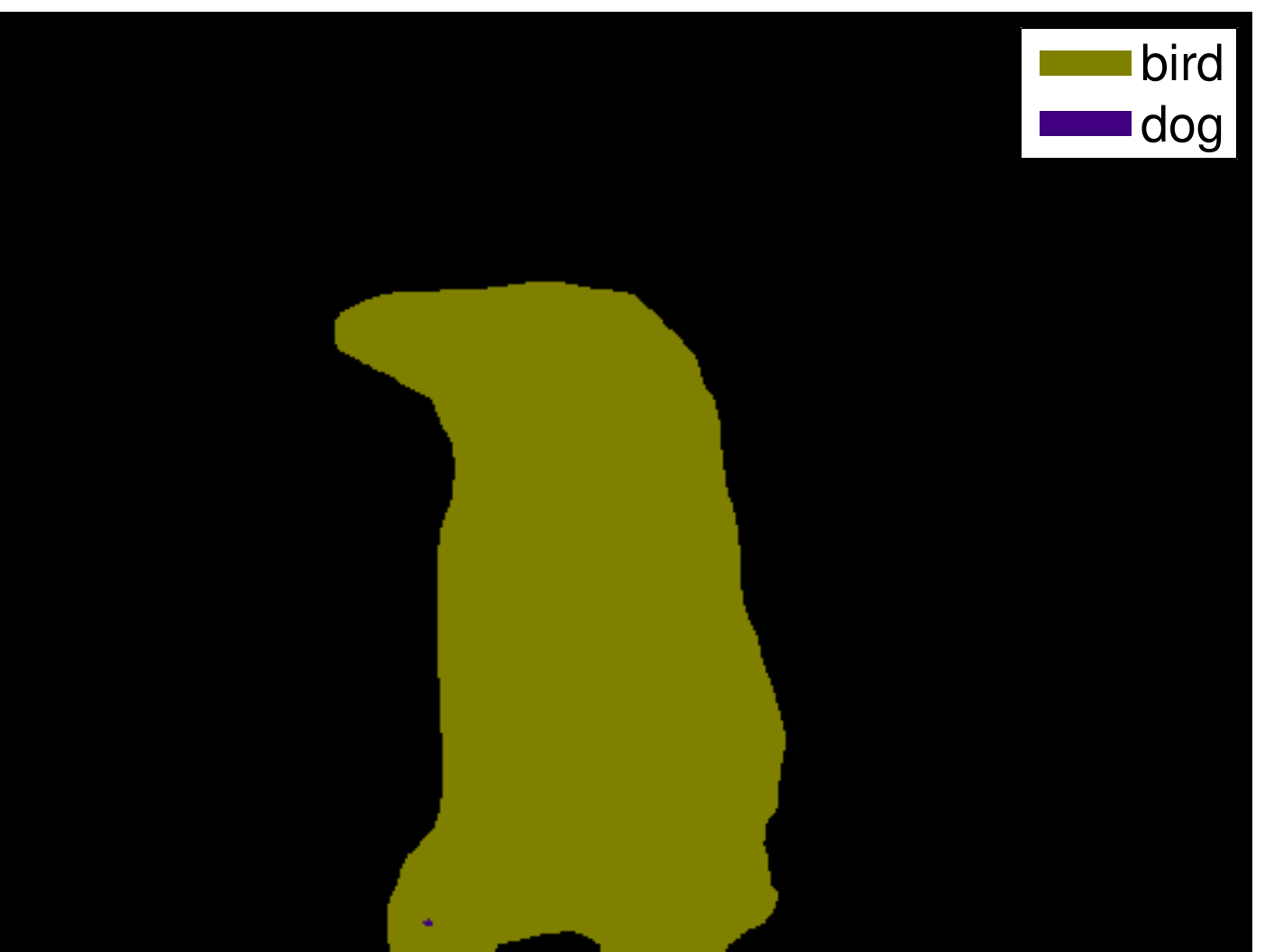}\\
	\begin{subtable}[b]{0.24\linewidth}
		\includegraphics[width=\linewidth]{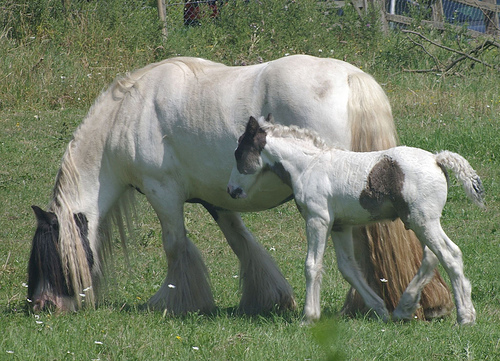}
		\caption{Original Image}
	\end{subtable}
	\begin{subtable}[b]{0.24\linewidth}
		\includegraphics[width=\linewidth]{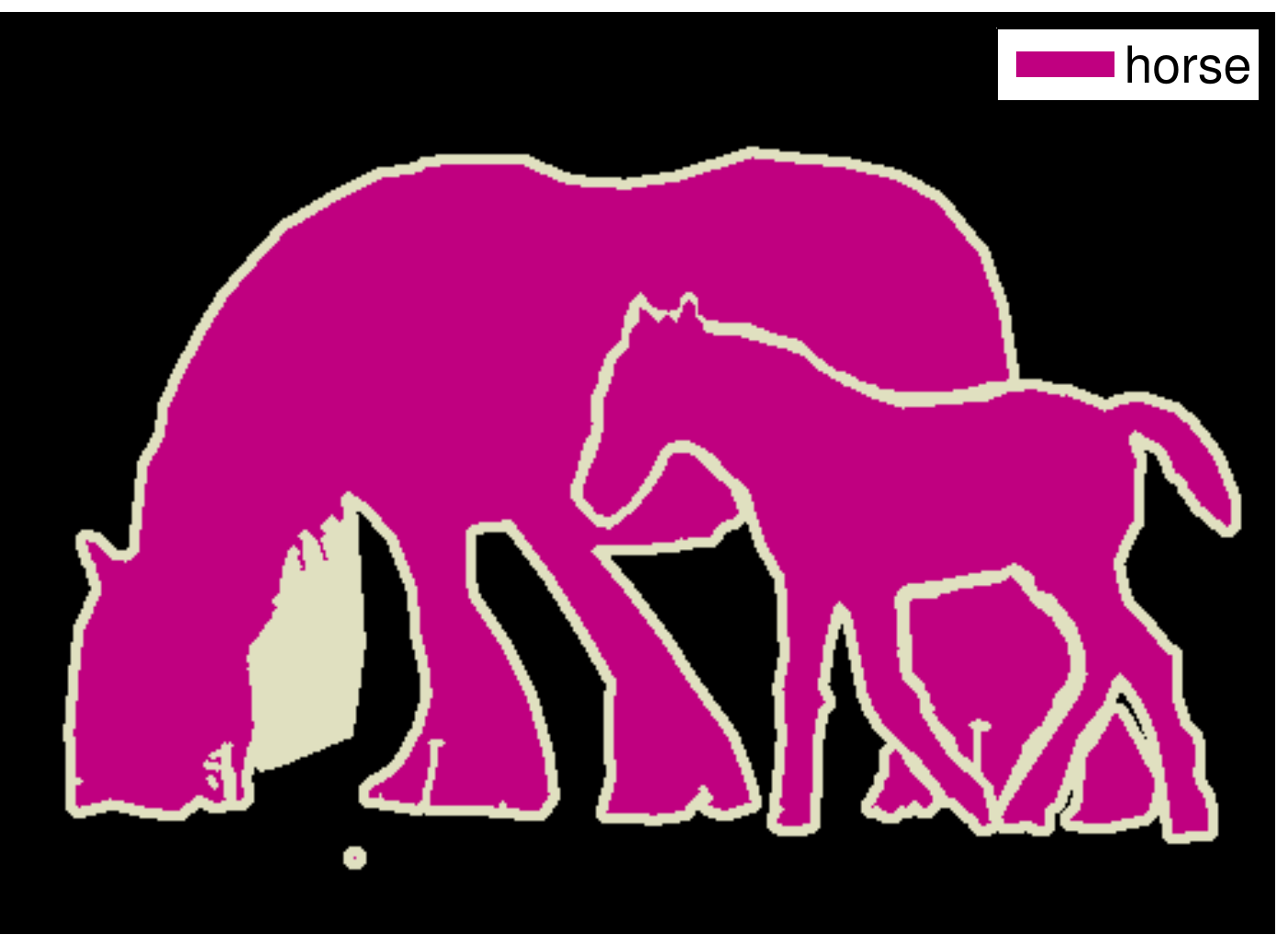}
		\caption{Ground truth}
	\end{subtable}
	\begin{subtable}[b]{0.24\linewidth}
		\includegraphics[width=\linewidth]{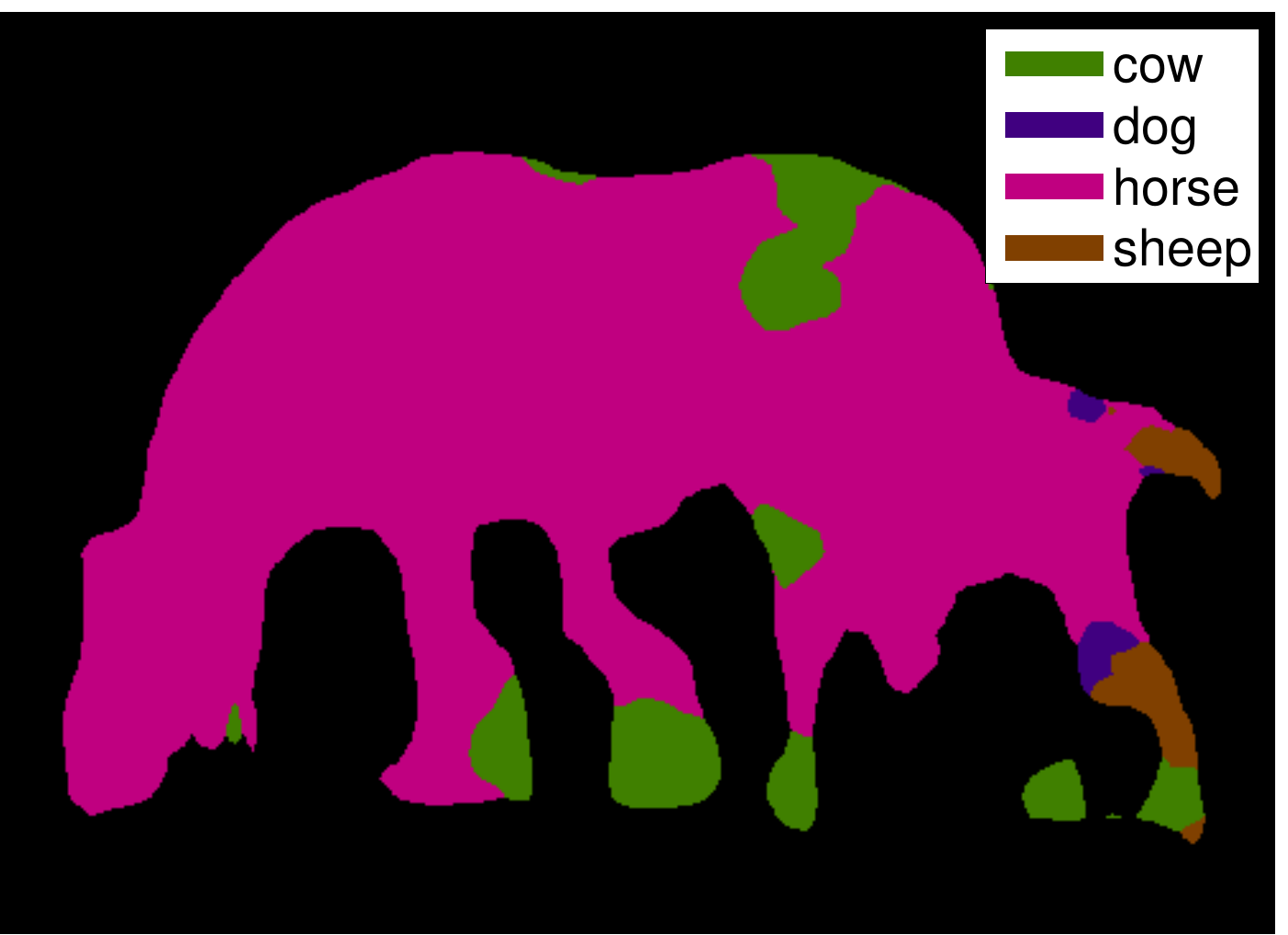}
		\caption{ParseNet Baseline}
	\end{subtable}
	\begin{subtable}[b]{0.24\linewidth}
		\includegraphics[width=\linewidth]{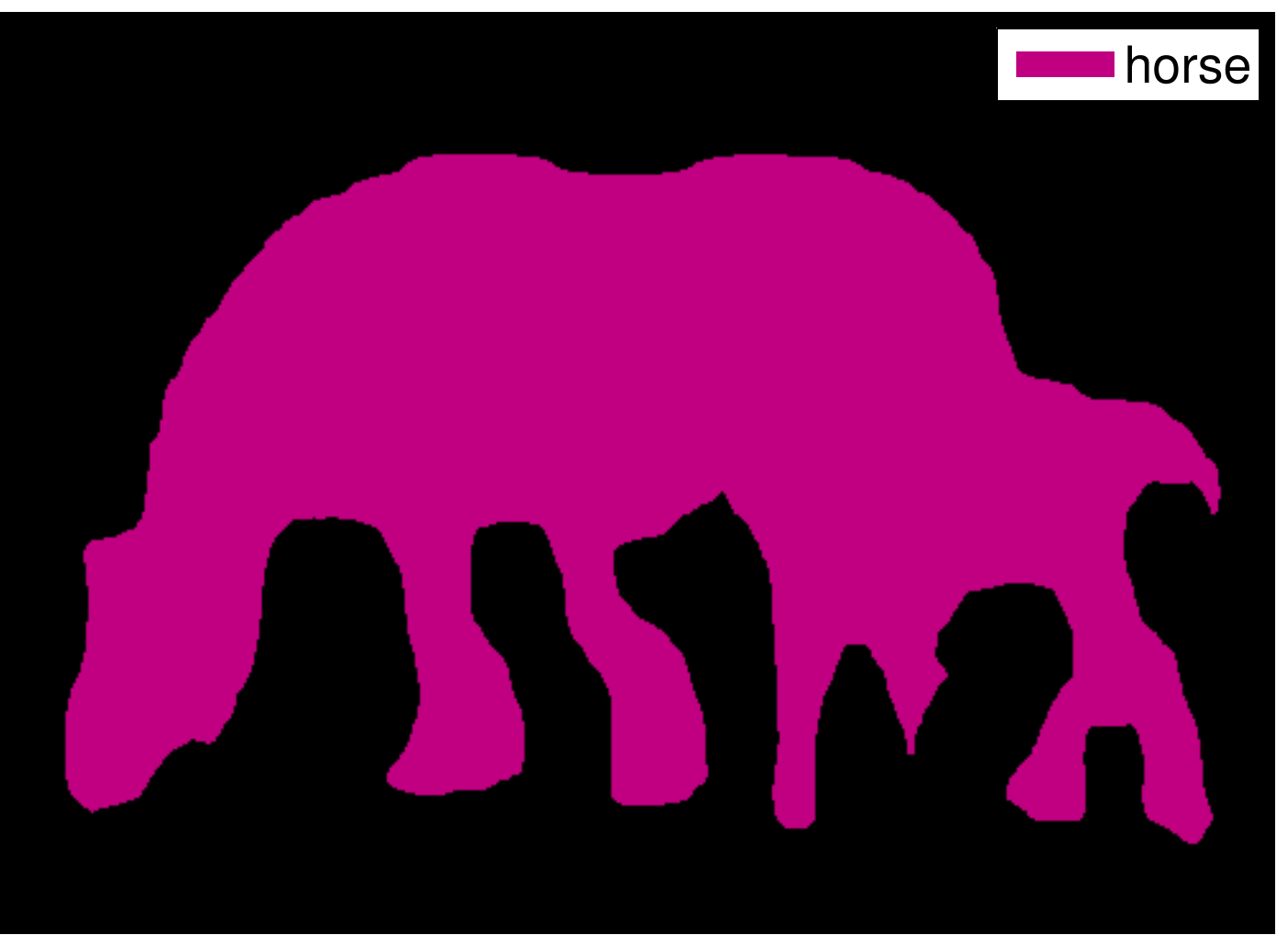}
		\caption{ParseNet}
	\end{subtable}
	\caption{Global context helps for classifying local patches.}
	\label{fig:globalcontexthelps}
\end{figure}

\begin{figure}
	\centering
	\includegraphics[width=0.24\linewidth]{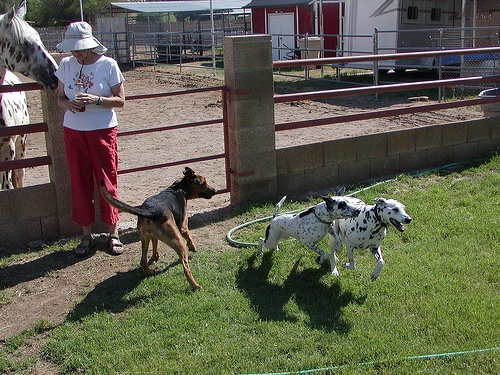}
	\includegraphics[width=0.24\linewidth]{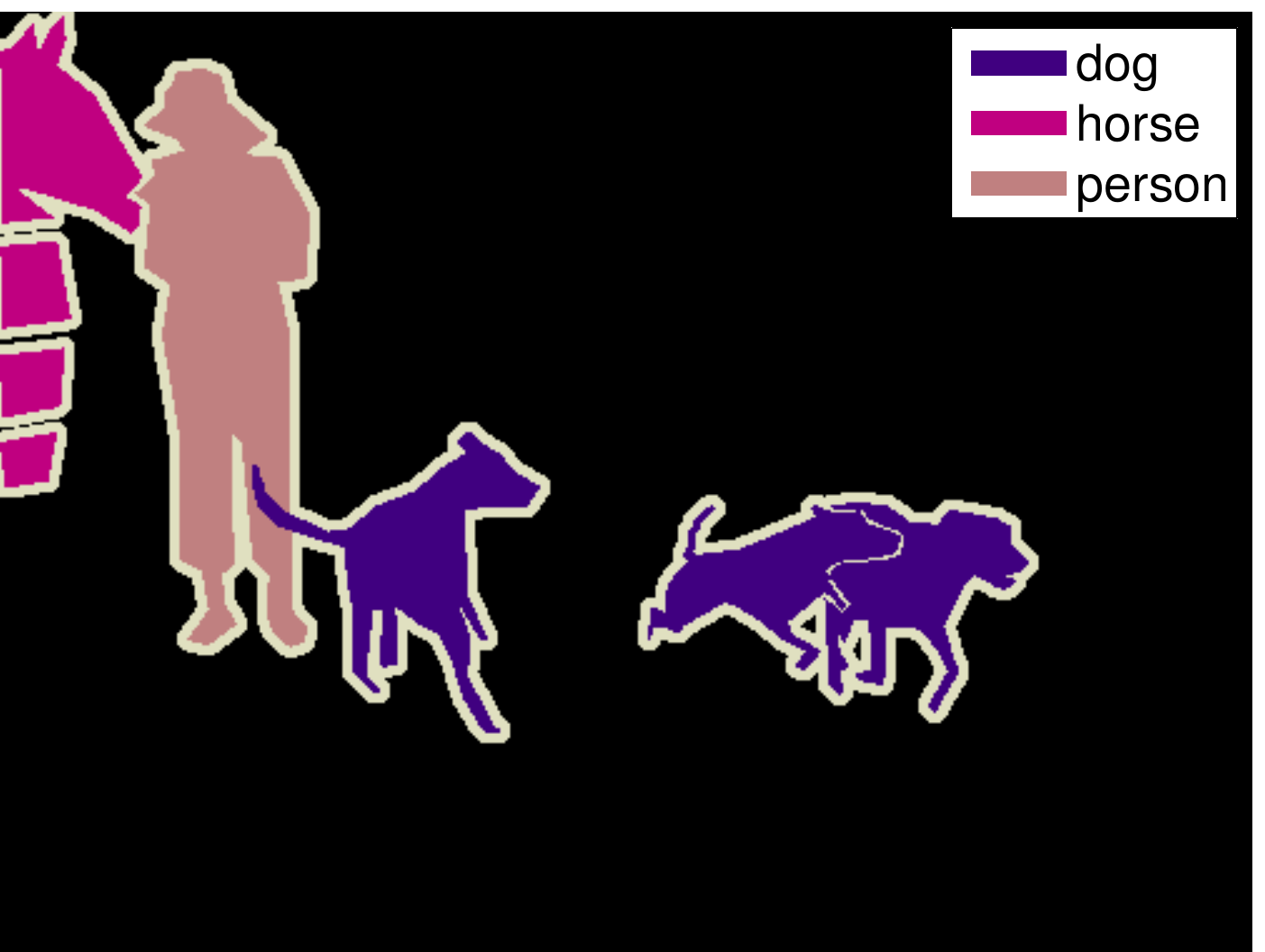}
	\includegraphics[width=0.24\linewidth]{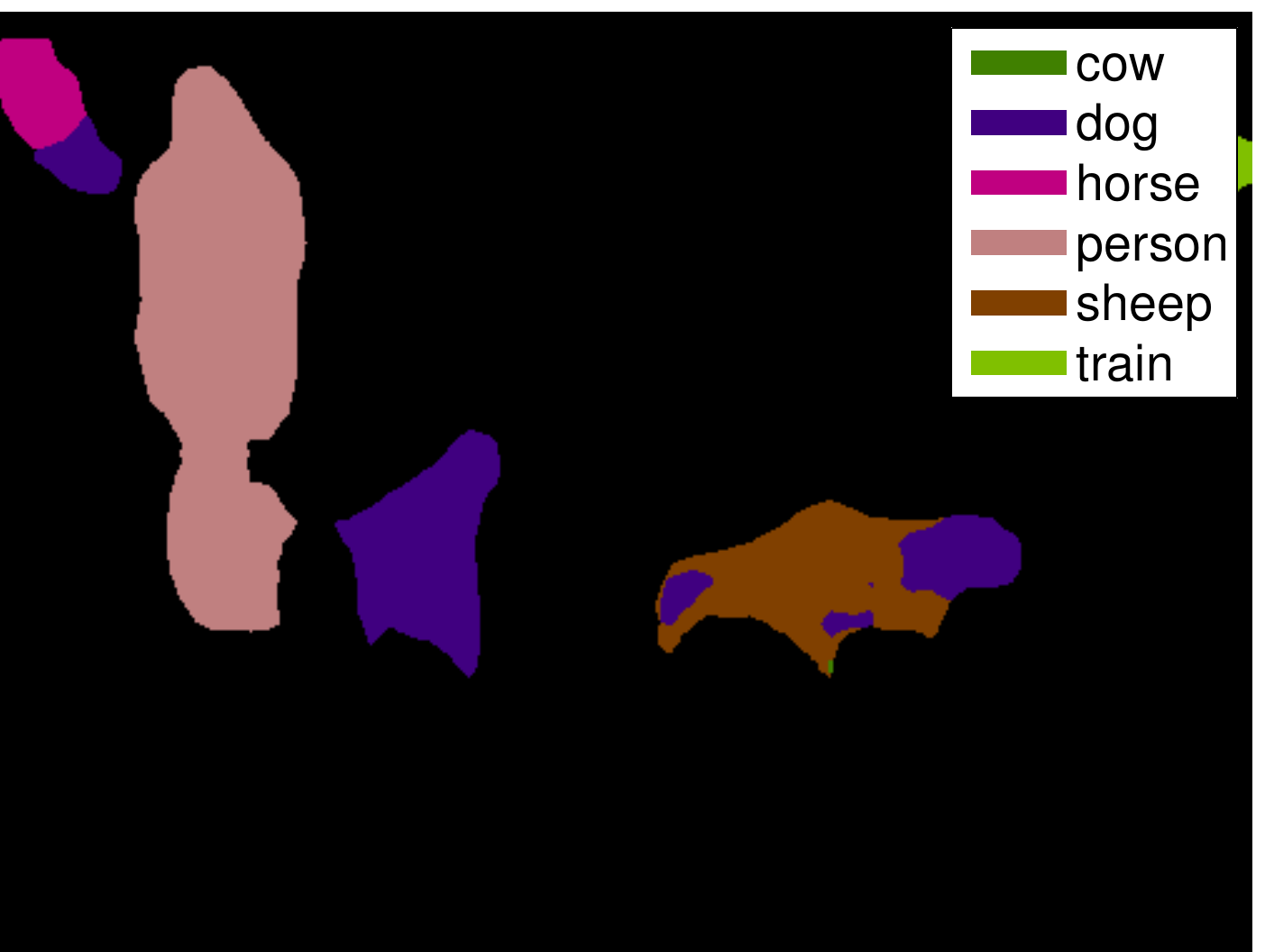}
	\includegraphics[width=0.24\linewidth]{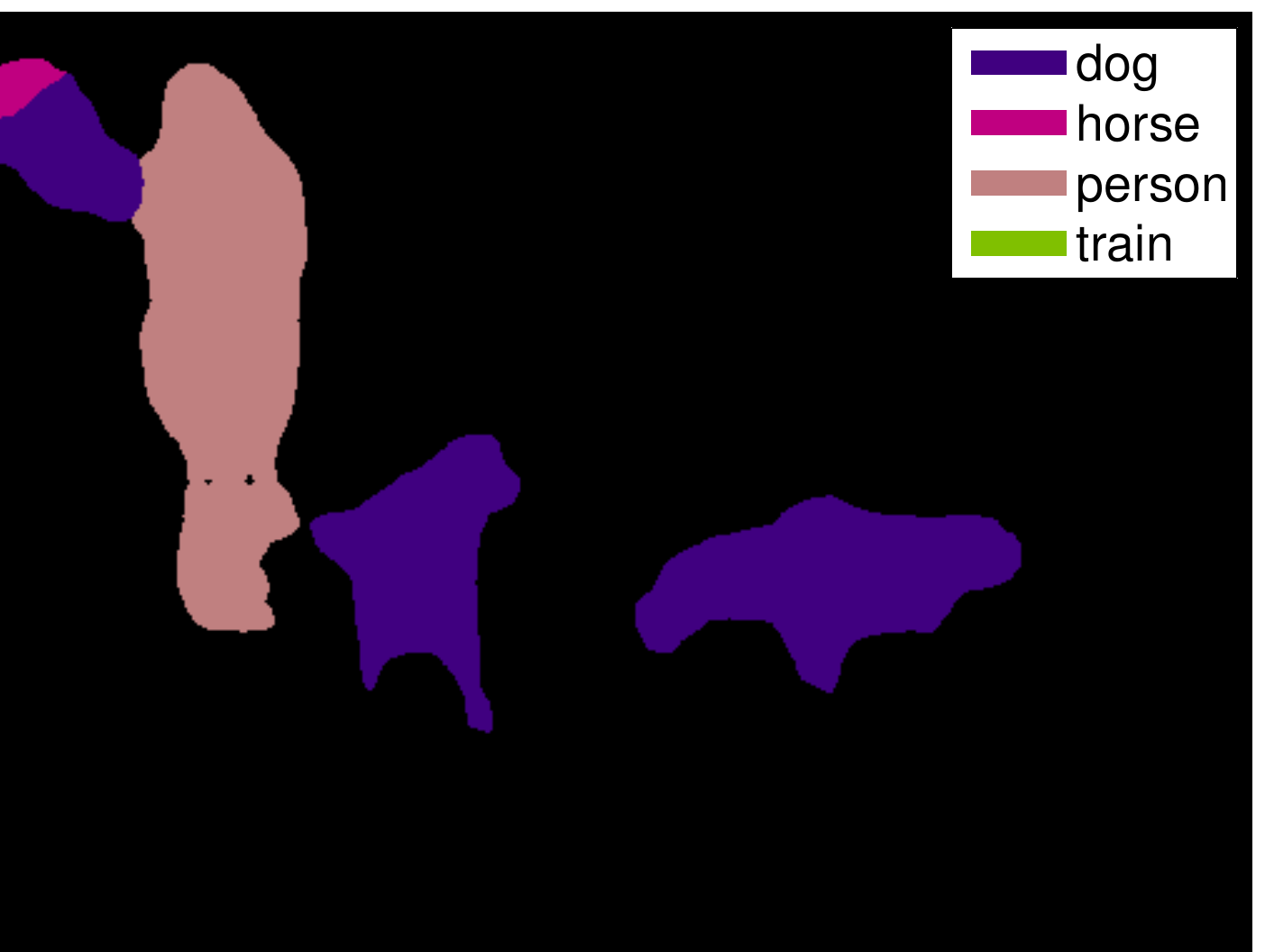}\\
	\includegraphics[width=0.24\linewidth]{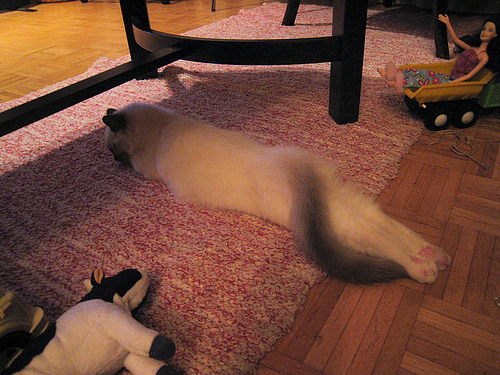}
	\includegraphics[width=0.24\linewidth]{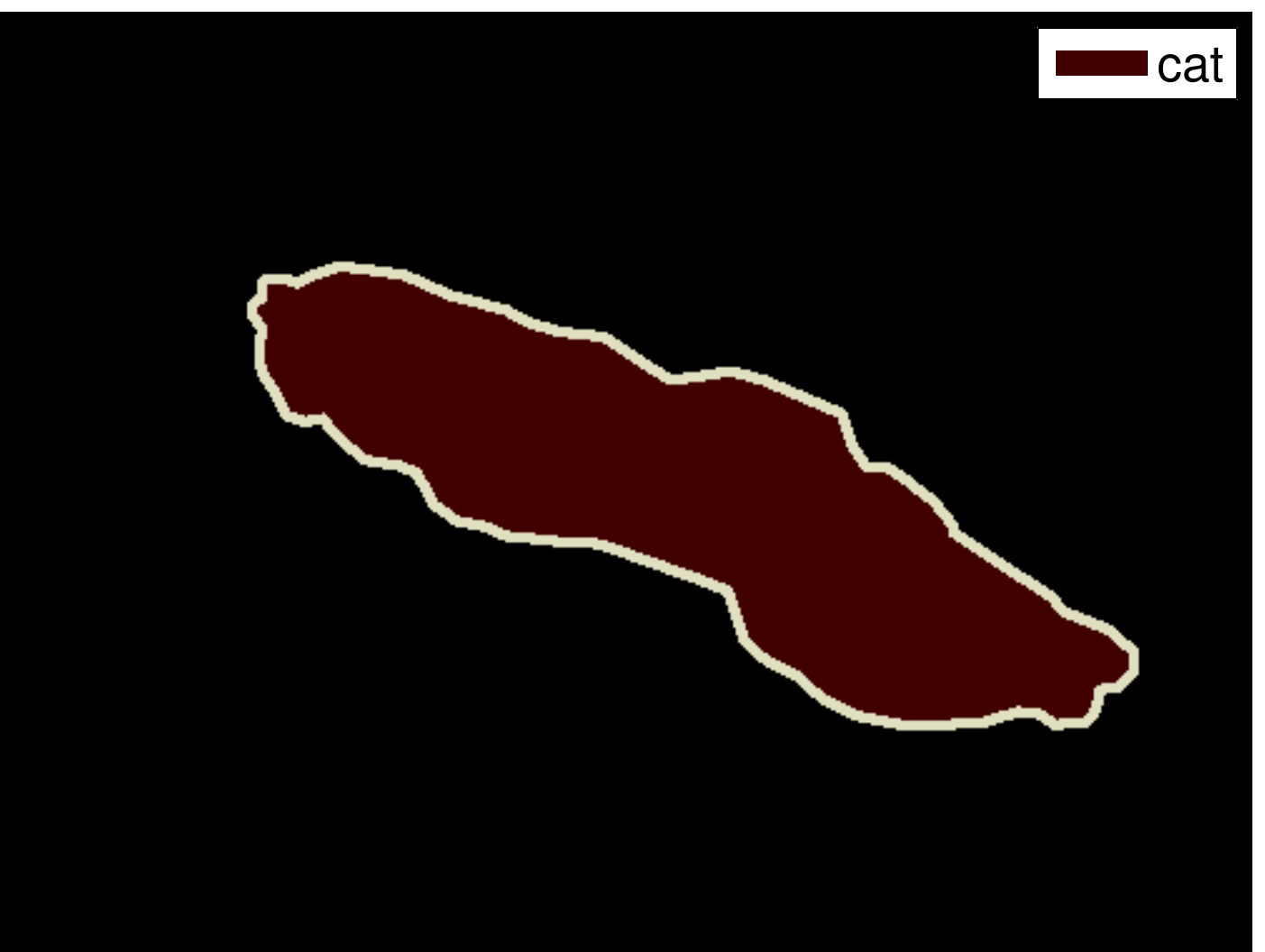}
	\includegraphics[width=0.24\linewidth]{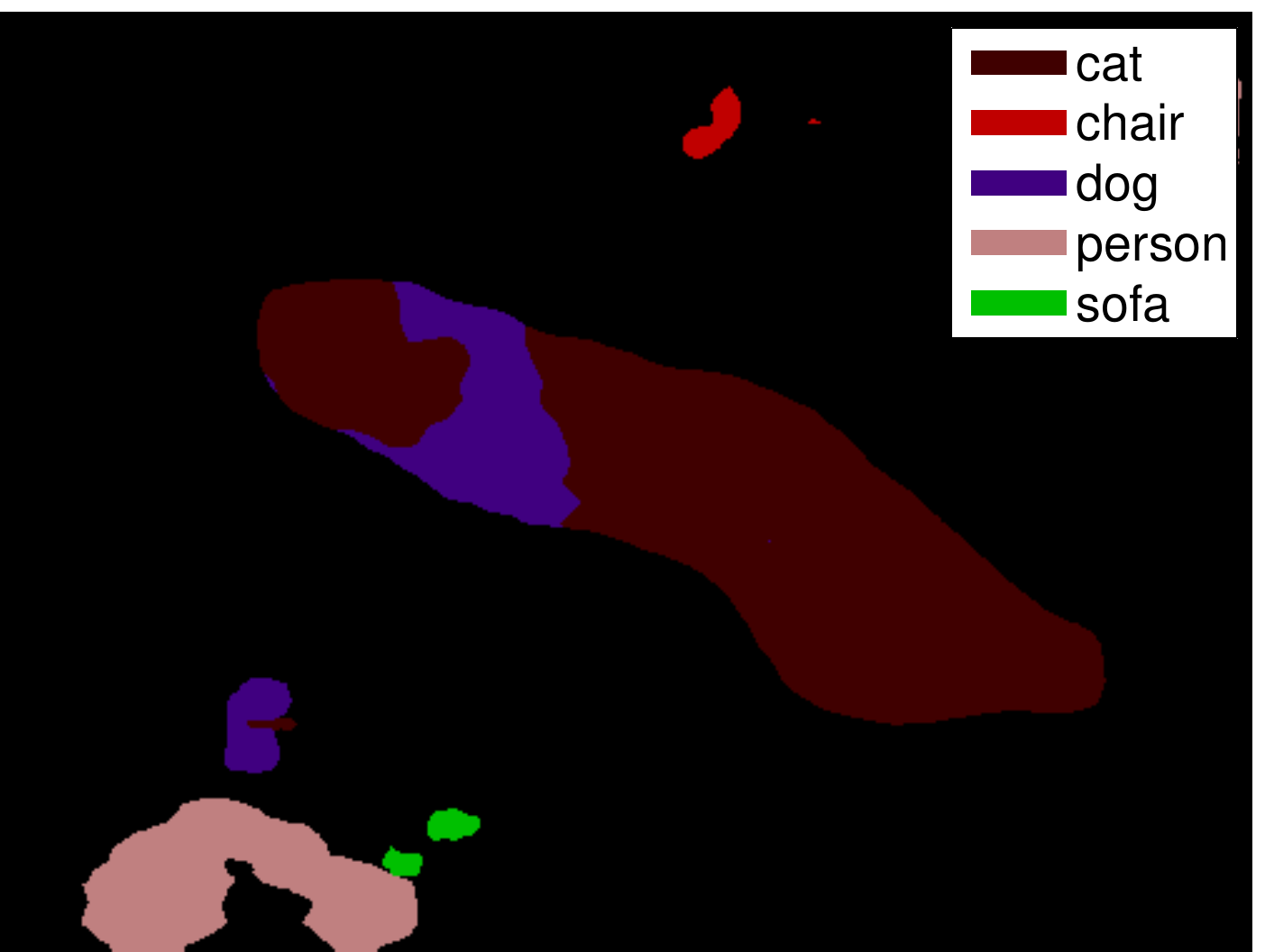}
	\includegraphics[width=0.24\linewidth]{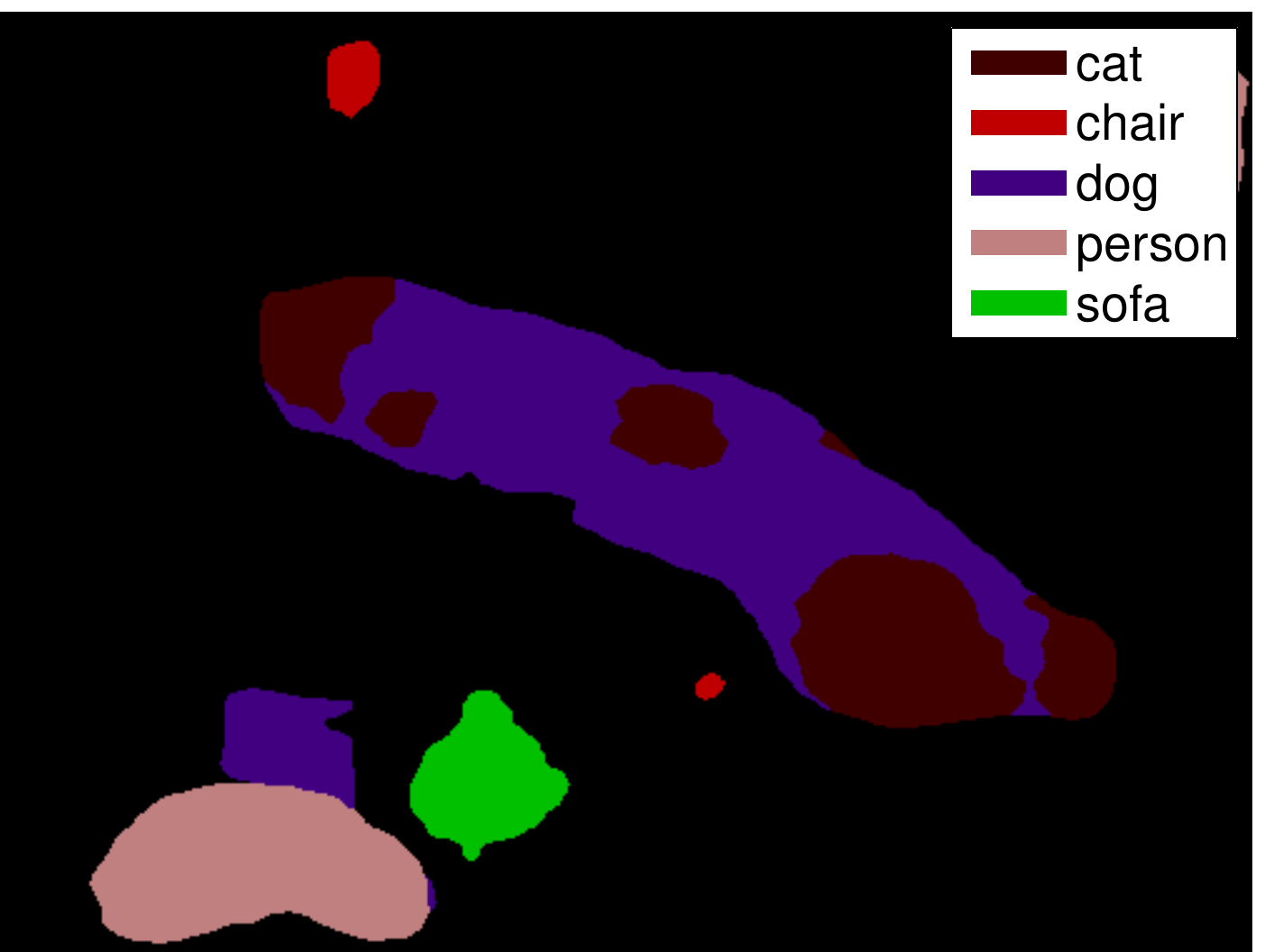}\\
	\includegraphics[width=0.24\linewidth]{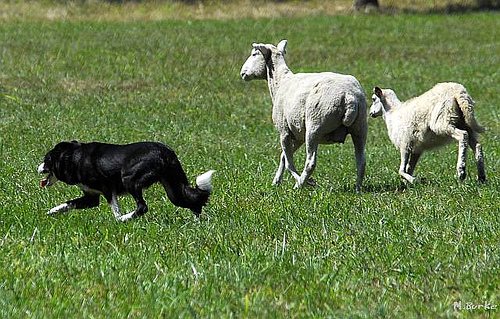}
	\includegraphics[width=0.24\linewidth]{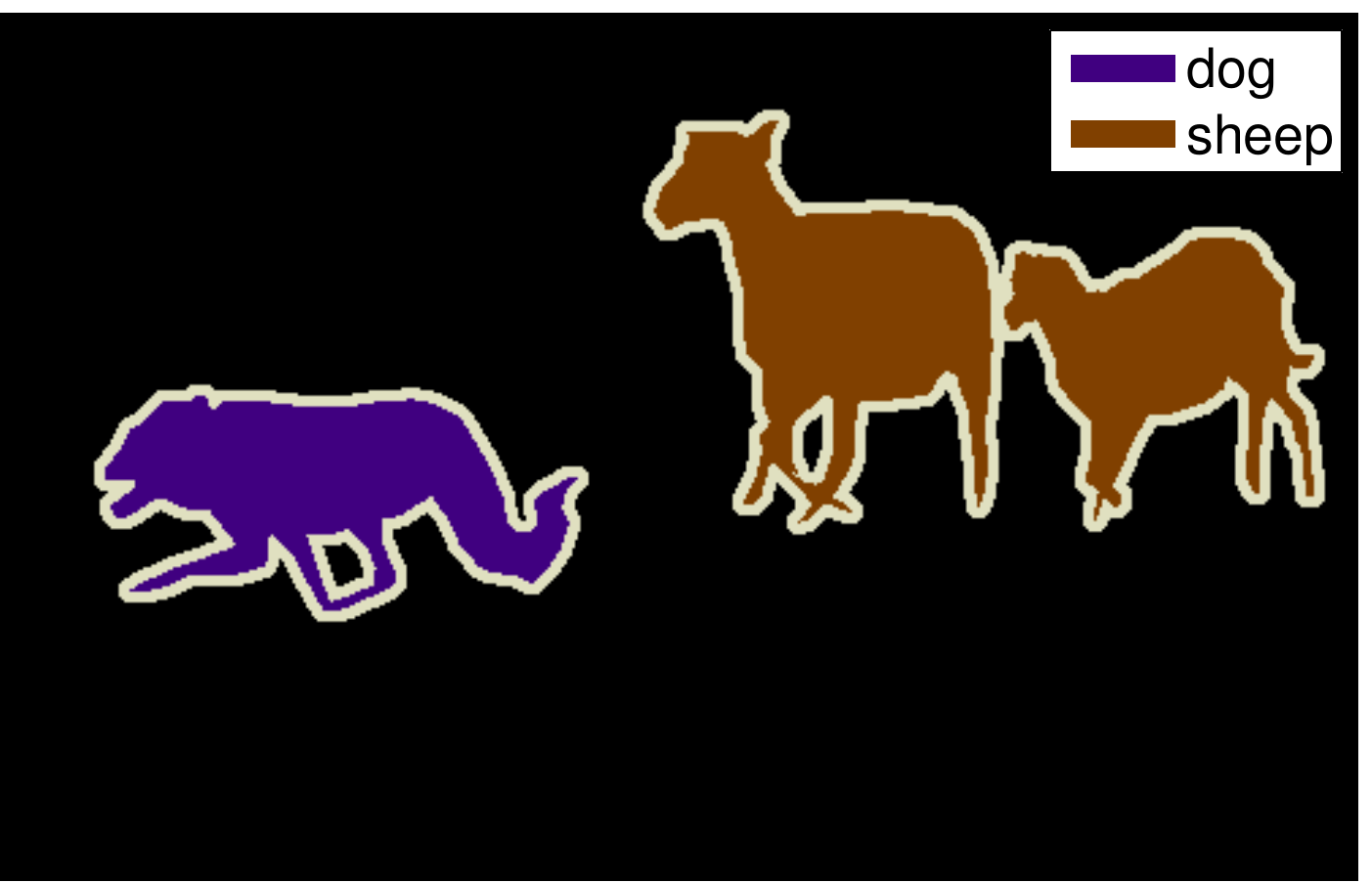}
	\includegraphics[width=0.24\linewidth]{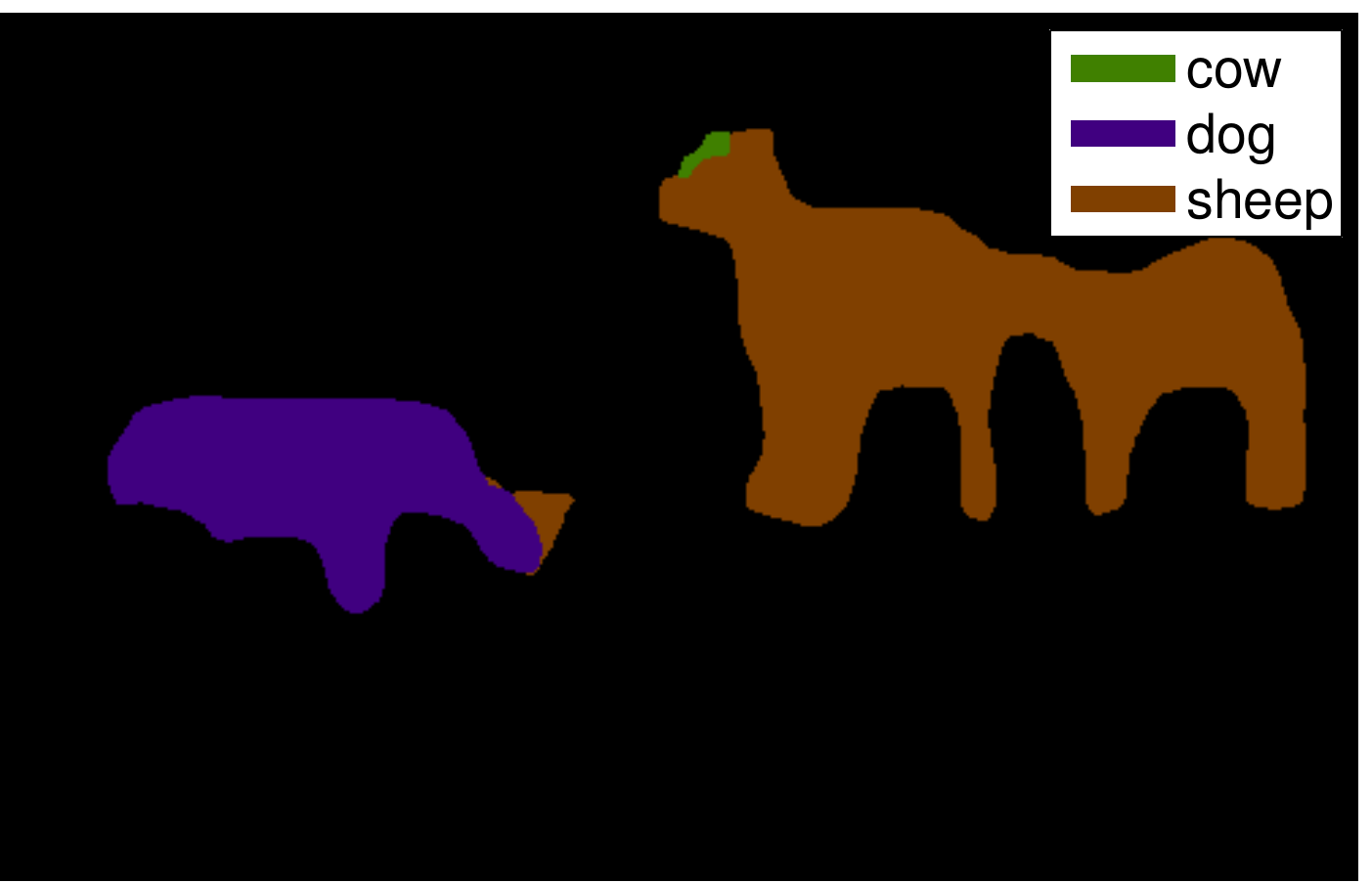}
	\includegraphics[width=0.24\linewidth]{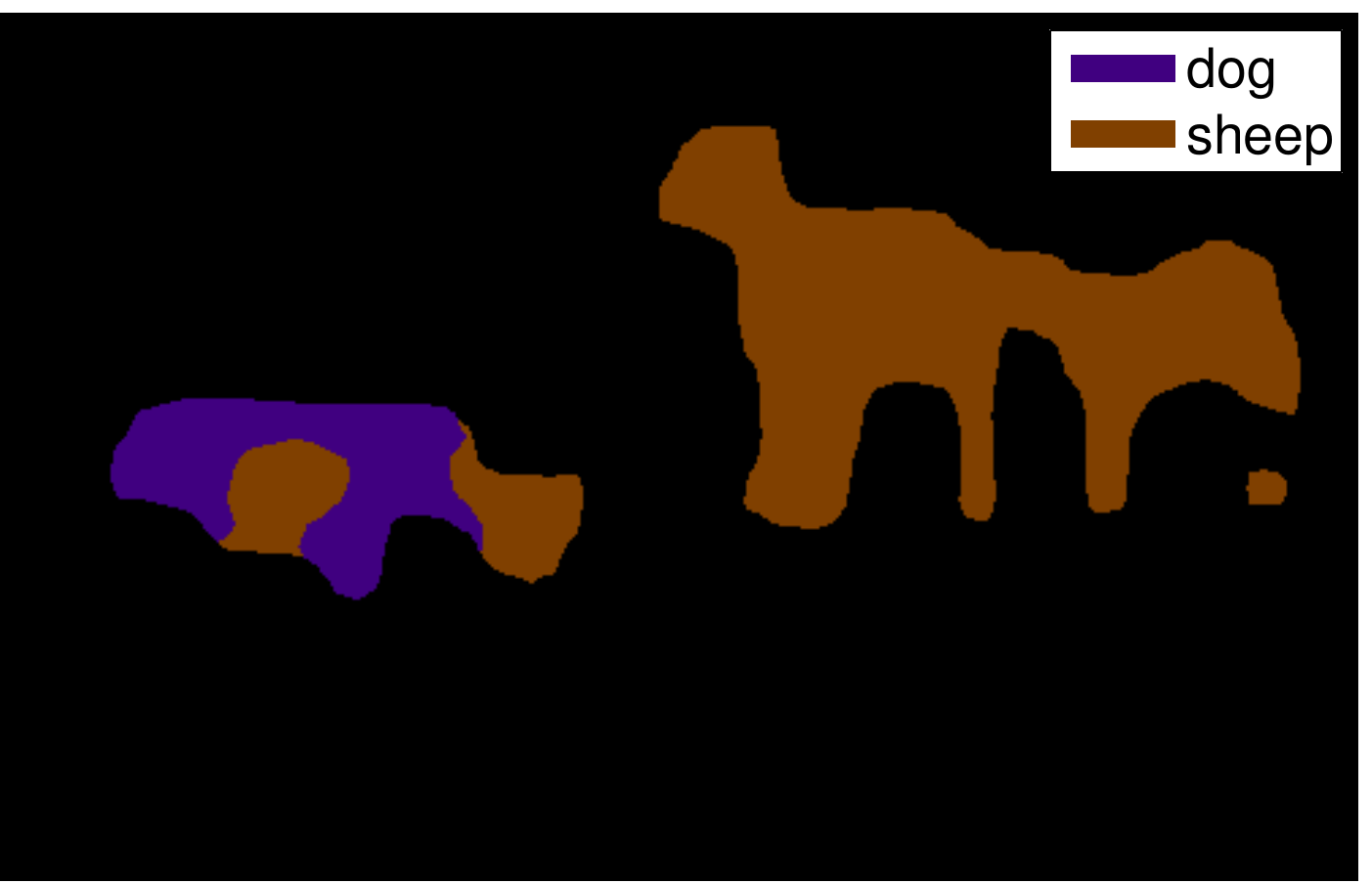}\\
	\begin{subtable}[b]{0.24\linewidth}
		\includegraphics[width=\linewidth]{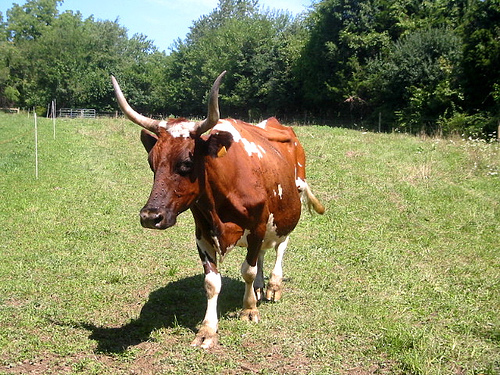}
		\caption{Original Image}
	\end{subtable}
	\begin{subtable}[b]{0.24\linewidth}
		\includegraphics[width=\linewidth]{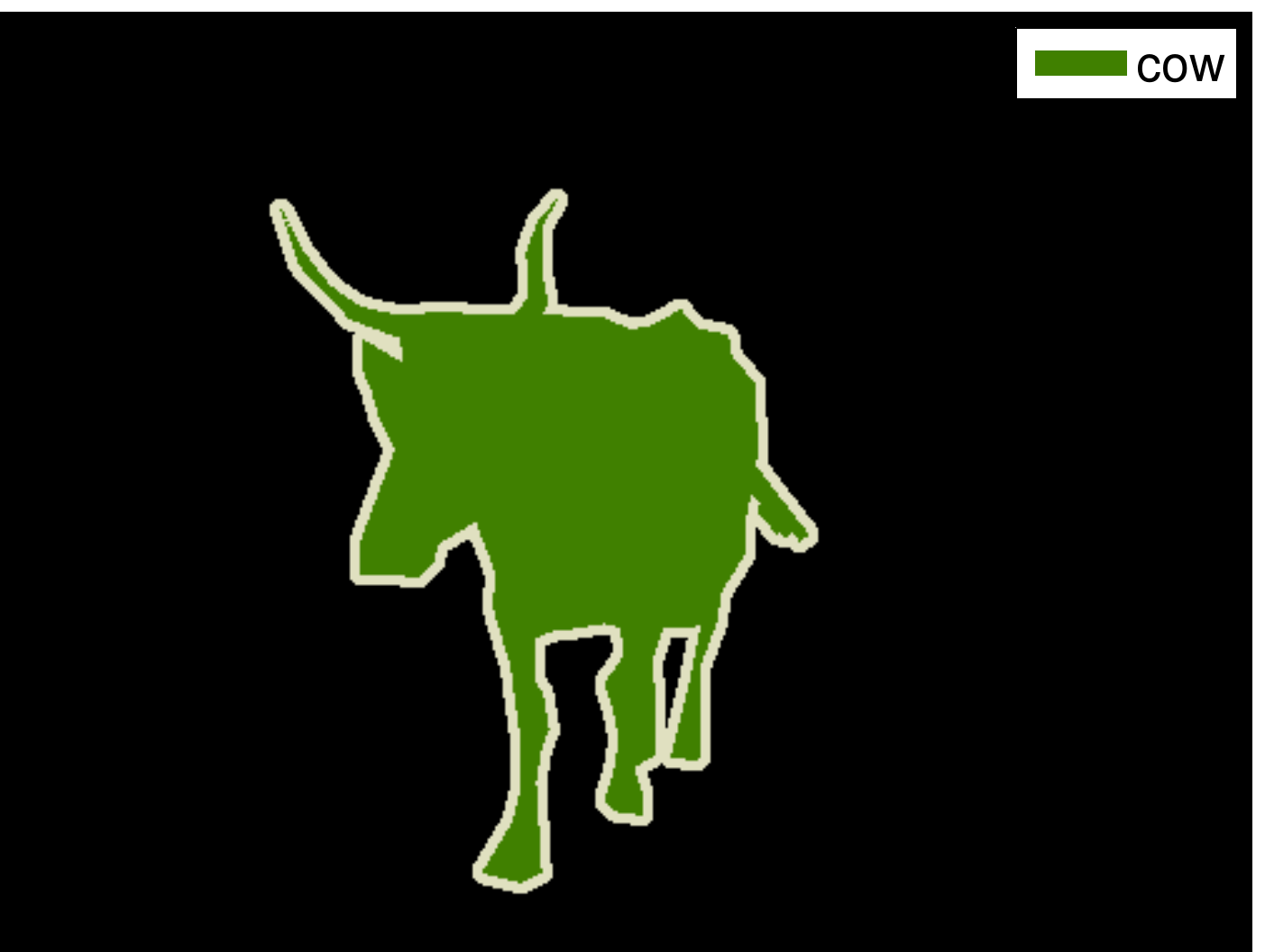}
		\caption{Ground truth}
	\end{subtable}
	\begin{subtable}[b]{0.24\linewidth}
		\includegraphics[width=\linewidth]{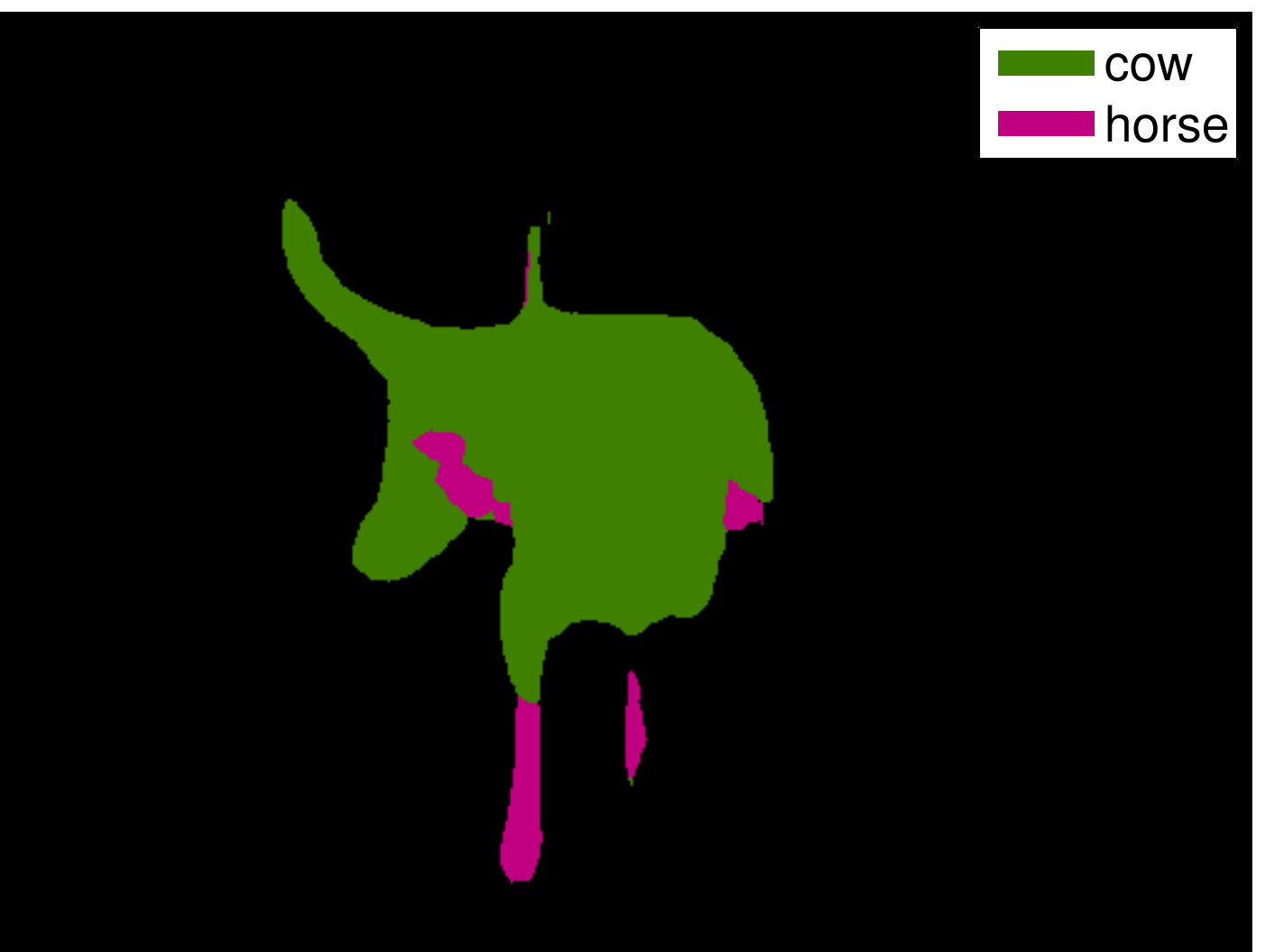}
		\caption{ParseNet Baseline}
	\end{subtable}
	\begin{subtable}[b]{0.24\linewidth}
		\includegraphics[width=\linewidth]{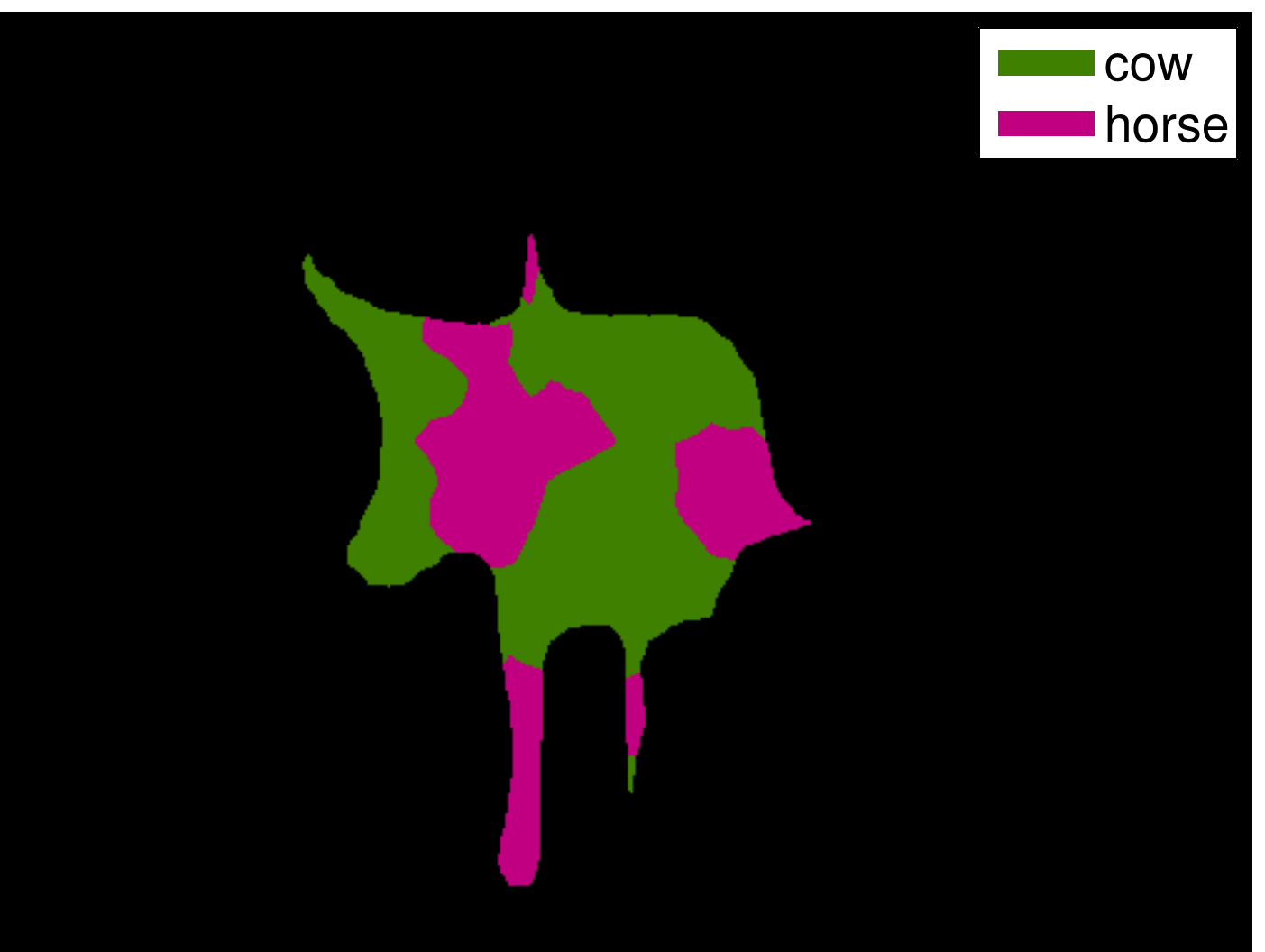}
		\caption{ParseNet}
	\end{subtable}
	\caption{Global context confuse local patch predictions.}
	\label{fig:globalcontextconfuse}
\end{figure}

ParseNet Baseline performance is higher than DeepLab Baseline and it is faster, thus we switch to use it for most of the experimental comparison for VOC2012. As shown in Table~\ref{tab:voc2012fov}, we observe a similar pattern as of DeepLab Baseline that if we add pool6, it is helping improve the performance by 3.8\%. However, we also notice that if we do not normalize them and learn the scaling factors, its effect is diminished. Furthermore, we notice that early fusion and late fusion both work very similar. Figure~\ref{fig:globalcontexthelps} illustrates some examples of how global context helps. We can clearly see that without using context feature, the network will make many mistakes by confusing between similar categories as well as making spurious predictions. Two similar looking patches are indistinguishable by the network if considered in isolation. However, adding context solves this issue as the global context helps discriminate the local patches more accurately. On the other hand, sometimes context also brings confusion for prediction as shown in Figure~\ref{fig:globalcontextconfuse}. For example, in the first row, the global context feature definitely captured the spotty dog information that it used to help discriminate sheep from dog. However, it also added bias to classify the spotty horse as a dog. The other three examples have the same issue. Overall, by learning to weight pool6 and fc7 after $L_2$ normalization helps improve the performance greatly.
\begin{table}[!htb]
	\centering
	\begin{tabular}{c|c|m{4em}|c}
		Layers & Norm (Y/N) & Early or Late (E/L) & Mean IoU\\
		\hline\hline
		fc7 & N & NA & 65.82\\
		fc7 & Y & NA & 65.66\\
		pool6 + fc7 & N & E & 65.30\\
		pool6 + fc7 & Y & E & 69.43\\
		pool6 + fc7 & Y & L & \textbf{69.55}\\
		pool6 + fc7 & N & L & 69.29\\
		\hline
	\end{tabular}
	\caption{\textbf{Add context for ParseNet Baseline on VOC2012.}}\label{tab:voc2012fov}
\end{table}
\vspace{-0.5em}

We also tried to combine lower level feature as was done with PASCAL-Context and SiftFlow, but no significant improvements using either \emph{early fusion} or \emph{late fusion} were observed. We believe it is because the fc7 of ParseNet Baseline is the same size as of conv4, and including lower level feature will not help much as they are not sufficiently discriminative. Besides, we also tried the idea similar to spatial pyramid pooling where we pool $1\times 1$ global feature, $2\times 2$ subregion feature, and $4\times 4$ subregion feature, and tried both \emph{early fusion} and \emph{late fusion}. However, we observed no improvements. We conjecture that the receptive field of the high level feature map (e.g. fc7) is sufficiently large that sub-region global feature does not help much.

\begin{savenotes}
\begin{table}[htbp]\tiny
	\centering
	\setlength{\tabcolsep}{1.6pt}
	\begin{tabular*}{\textwidth}{l@{\hspace{0.1cm}}cccccccccccccccccccccc}
		\toprule
		\noalign{\smallskip}
		\textbf{System} & \textbf{bkg} & \textbf{aero} & \textbf{bike} & \textbf{bird} & \textbf{boat} & \textbf{bottle} & \textbf{bus} & \textbf{car} & \textbf{cat} & \textbf{chair} & \textbf{cow} & \textbf{table} & \textbf{dog} & \textbf{horse} & \textbf{mbike} & \textbf{person} & \textbf{plant} & \textbf{sheep} & \textbf{sofa} & \textbf{train} & \textbf{tv} & \emph{mean}\\ \noalign{\smallskip}\cline{2-23}
		\noalign{\smallskip}
		FCN-8s & - & 76.8 & 34.2 & 68.9 & 49.4 & 60.3 & 75.3 & 74.7 & 77.6 & 21.4 & 62.5 & 46.8 & 71.8 & 63.9 & 76.5 & 73.9 & 45.2 & 72.4 & 37.4 & 70.9 & 55.1 & 62.2\\
		Hypercolumn & - & 68.7 & 33.5 & 69.8 & 51.3 & 70.2 & 81.1 & 71.9 & 74.9 & 23.9 & 60.6 & 46.9 & 72.1 & 68.3 & 74.5 & 72.9 & 52.6 & 64.4 & 45.4 & 64.9 & 57.4 & 62.6\\
		TTI-Zoomout-16 & 89.8 & 81.9 & 35.1 & 78.2 & 57.4 & 56.5 & 80.5 & 74.0 & 79.8 & 22.4 & 69.6 & 53.7 & 74.0 & 76.0 & 76.6 & 68.8 & 44.3 & 70.2 & 40.2 & 68.9 & 55.3 & 64.4\\
		\scriptsize{DeepLab-LargeFOV} & \textbf{92.6} & 83.5 & 36.6 & \textbf{82.5} & 62.3 & \textbf{66.5} & 85.4 & 78.5 & 83.7 & \textbf{30.4} & 72.9 & \textbf{60.4} & \textbf{78.5} & 75.5 & \textbf{82.1} & \textbf{79.7} & \textbf{58.2} & \textbf{82.0} & 48.8 & 73.7 & 63.3 & \textbf{70.3}\\
		\hline\hline
		ParseNet Baseline ~\footnote{\url{http://host.robots.ox.ac.uk:8080/anonymous/LGOLRG.html}} & 92.3 & 82.6 & 36.1 & 76.1 & 59.3 & 62.3 & 81.6 & 79.5 & 81.4 & 28.1 & 70.0 & 53.0 & 73.2 & 70.6 & 78.8 & 78.6 & 51.9 & 77.4 & 45.5 & 71.7 & 62.6 & 67.3\\
        ParseNet ~\footnote{\url{http://host.robots.ox.ac.uk:8080/anonymous/56QLXU.html}} & 92.4 & \textbf{84.1} & \textbf{37.0} & 77.0 & \textbf{62.8} & 64.0 & \textbf{85.8} & \textbf{79.7} & \textbf{83.7} & 27.7 & \textbf{74.8} & 57.6 & 77.1 & \textbf{78.3} & 81.0 & 78.2 & 52.6 & 80.4 & \textbf{49.9} & \textbf{75.7} & \textbf{65.0} & 69.8\\
		\bottomrule
	\end{tabular*}
	\caption{PASCAL VOC2012 test Segmentation results.}\label{tab:pascalseg}
\end{table}
\end{savenotes}
Finally, we test two models, ParseNet Baseline and ParseNet, on VOC2012 test set. As shown in Table~\ref{tab:pascalseg}, we can see that our baseline result is already higher than many of the existing methods due to proper finetuning. By adding the global context feature, we achieve performance that is within the standard deviation of the one~\cite{chen2014semantic} using fully connect CRF to smooth the outputs and perform better on more than half of categories. Again, our approach is much simpler to implement and train, hence is more robust. Using \emph{late fusion} has almost no extra training/inference cost.

\captionsetup[subfigure]{singlelinecheck=off,justification=raggedright}
\section{Conclusion}
\label{sec:futurework}
In this work we presented ParseNet, a simple fully convolutional neural network architecture that allows for direct inclusion of global context for the task of semantic segmentation. We have explicitly demonstrated that relying on the largest receptive field of FCN network does not provide sufficient global context, and the largest empirical receptive field is not sufficient to capture global context -- modeling global context directly in required. On PASCAL VOC2012 test set, segmentation results of ParseNet are within the standard deviation of the DeepLab-LargeFOV-CRF, which suggests that adding a global feature has a similar effect of post processing FCN predictions with a graphical model. As part of developing and analyzing this approach we provided analysis of many architectural choices for the network, discussing best practices for training, and demonstrated the importance of normalization and learning weights when combining features from multiple layers of a network. By themselves, our practices for training significantly improve the baselines we use before adding global context. The guiding principle in the design of ParseNet is simplicity and robustness of learning. Results are presented on three benchmark dataset, and are state of the art on SiftFlow and PASCAL-Context, and near the state of the art on PASCAL VOC2012. Given the simplicity and ease of training, we find these results very encouraging. In our on going work, we are exploring combining our technique with structure training/inference as done in~\cite{schwing2015fully, lin2015efficient, zheng2015conditional}.
\bibliography{parsenet}
\bibliographystyle{iclr2016_conference}

\end{document}